\documentclass{article}

\usepackage{microtype}
\usepackage{amssymb,amsmath,amsthm,dsfont,bm,mathtools}
\usepackage{color,graphicx,epsfig,colortbl}
\usepackage{pdfpages}
\usepackage{bigstrut} %multirow,rotating}
\usepackage{capt-of,caption,subfig}
\usepackage{wrapfig}
\usepackage[title]{appendix}
\usepackage[multiple]{footmisc}
\usepackage{tikz}  

\usepackage{hyperref}

\usepackage[capitalise]{cleveref}

\usepackage{epigraph}

\newcommand{\tblfontsz}{\fontsize{8.5pt}{9.5pt}\selectfont}
\newcommand{\algofontsz}{}

\newcommand{\loss}{{\ell}}
\newcommand{\des}{{a}}
\newcommand{\Des}{{\A}}
\newcommand{\yhat}{{\hat{y}}}
\newcommand{\ytilde}{{\tilde{y}}}
\newcommand{\ybar}{{\bar{y}}}
\newcommand{\predm}{{\ytilde}}

\newcommand{\hhat}{{\hat{h}}}
\newcommand{\sidehat}{{\hat{\side}}}

\newcommand{\smplst}{{\mathcal{S}}}
\newcommand{\smplstaux}{{\mathcal{T}}}
\newcommand{\reg}{{R}}

\newcommand{\nn}{{\mathcal{N}}}
\newcommand{\repnet}{{\phi}}

\newcommand{\repprm}{{\theta}}
\newcommand{\Repprm}{{\Theta}}
\newcommand{\hhatprm}{{\eta}}
\newcommand{\Hhatprm}{{H}}

\newcommand{\viz}{{v}}
\newcommand{\vizmap}{{\rho}}
\newcommand{\vizobjs}{{\mathcal{V}}}
\newcommand{\decoder}{{\psi}}
\newcommand{\advice}{{\gamma}}
\newcommand{\Advice}{{\Gamma}}
\newcommand{\side}{{s}}

\newcommand{\zdim}{{9}}

\newcommand{\mom}{{M\raisebox{0.6pt}{$\circ$}M}}
\newcommand{\feat}[1]{{\textsc{{#1}}}}

\usepackage{xcolor}

\newcount\Comments  % 0 suppresses notes to selves in text

\newcommand{\kibitz}[2]{\ifnum\Comments=1{\color{#1}{#2}}\fi}
\newcommand{\dcp}[1]{\kibitz{blue}{[DCP: #1]}}

\newcommand{\sophie}[1]{\kibitz{orange}{[SOPHIE: #1]}}

\newcount\CommentsNew  % leave this at 1 , else we drop the things added
\newcommand{\kibitzNew}[2]{\ifnum\CommentsNew=1{\color{#1}{#2}}\fi}

\newcommand{\beq}{\begin{equation}}
\newcommand{\eeq}{\end{equation}}
\newcommand{\bal}{\begin{align}}
\newcommand{\eal}{\end{align}}
\newcommand\expect[2]{\mathbb{E}_{#1}{[ {#2} ]}}

\DeclareMathOperator*{\argmin}{argmin}

\newcommand\inner[1]{\langle {#1} \rangle}

\newcommand{\1}[1]{\mathds{1}{\{{#1}\}}}
\newcommand{\one}[1]{\mathds{1}_{\{{#1}\}}}
\newcommand{\X}{{\cal{X}}}

\newcommand{\Y}{{\cal{Y}}}
\newcommand{\A}{{\cal{A}}}

\newcommand{\R}{{\mathbb{R}}}
\usepackage[accepted]{icml2021}

\icmltitlerunning{Learning Representations by Humans, for Humans}

\begin{document}

\twocolumn[
\icmltitle{Learning Representations by Humans, for Humans}



\icmlsetsymbol{equal}{*}

\begin{icmlauthorlist}
\icmlauthor{Sophie Hilgard}{seas,equal}
\icmlauthor{Nir Rosenfeld}{technion,equal}
\icmlauthor{Mahzarin Banaji}{psych}
\icmlauthor{Jack Cao}{psych}
\icmlauthor{David C. Parkes}{seas}
\end{icmlauthorlist}

\icmlaffiliation{seas}{School of Engineering and Applied Science, Harvard University, Cambridge, MA, USA}
\icmlaffiliation{technion}{Department of Computer Science, Technion - Israel Institute of Technology}
\icmlaffiliation{psych}{Department of Psychology, Harvard University, Cambridge, MA, USA}

\icmlcorrespondingauthor{Sophie Hilgard}{ash798@g.harvard.edu}
\icmlcorrespondingauthor{Nir Rosenfeld}{nirr@cs.technion.ac.il}

\icmlkeywords{Machine Learning, ICML}

\vskip 0.3in
]



\printAffiliationsAndNotice{\icmlEqualContribution} % otherwise use the standard text.

\begin{abstract}
% !TEX root = mom_neurips20.tex

% The task of optimizing machines to support human decision-making is often conflated with
% that of optimizing machines for accuracy even though they are materially different.
When machine predictors can achieve higher performance than the human decision-makers they support, improving 
the performance of human decision-makers is often conflated with improving
machine accuracy.
% the human decision problem 
% The efficacy of this solution hinges on a specific, simplified model of human decision-making, in which machine outputs are used as projected.
% Whereas it is typical for learning systems to prescribe actions through prediction,
% here we propose an approach in which the role of machines is to reframe problems,
% % in order to 
% directly 
% supporting human decision-making processes.
Here we propose a framework to directly support human decision-making, 
% using an approach 
in which the role of machines is to reframe problems rather than to prescribe actions through prediction.
Inspired by the success of representation learning in 
improving performance of machine predictors,
% transforming data to enhance the performance of machine predictors,
% we frame the problem as one of learning representations that are conducive to good human performance.
% we propose a flexible framework in which 
% machines seek to 
our framework learns human-facing representations 
% enable human decision-makers to achieve good performance.
optimized for human performance.
% are conducive to good human performance, while 
% taking into account the human decision process.
%that entail optimal human decisions.
This ``Mind Composed with Machine'' framework incorporates a human decision-making model directly into the 
representation learning paradigm
% with optimization achieved through 
and is trained with a novel human-in-the-loop training procedure.
We empirically demonstrate the
% \dcpadd{
successful application of the framework
% }
to  various tasks and representational forms.

\end{abstract}

% !TEX root = mom_neurips20.tex

% \todo{go over reviews, rebuttal, final responses and meta-review}

% \todo{add appendix tex file to main tex file}

\section{Introduction} \label{sec:intro}

\epigraph{\textit{``No one ever made a decision because of a number. \\ \hspace{0.1cm} They need a story.''}}{--- Daniel Kahneman}

%Across many important domains, a
Advancements in machine learning algorithms, as well as increased data availability and computational power, have led to the rise of predictive machines that outperform human experts in controlled experiments \cite{esteva2017dermatologist, nickerson2014political, tabibian2019enhancing}. 
However,
% is broad recognition that 
human involvement remains important in many domains, \cite{liu2019comparison},  
especially those in which safety and equity
are important considerations \cite{parikh2019regulation, barabas2017interventions}
and where users  have external information or want to exercise agency and
use their own judgment. In these settings, humans are the final arbiters, and the goal of algorithms is to produce useful decision aids.
%dcp cut space, and learning
%systems can  be brittle in the face of previously unseen situations. 
%
% As simple machine predictors are limited, so, too, are human arbiters:

% At the same time, humans 
% could clearly benefit from some aspects of machine prediction:
% % are limited in our capacity to make good decisions
% %  we are bound by 
%  our limited cognitive capabilities and 
%   vulnerabilities to psychological biases (related to, e.g., framing and heuristics)
% %   which
%  make it difficult to identify and rightly act upon
% complex patterns emerging in data \cite{kahneman2011thinking,miller1956magical}.

% To support decisions,
% and 
Given that learning algorithms excel at prediction,
previous efforts in this space
% to enable human decision-makers to employ data-driven machine predictions 
have largely focused on providing predictions as decision aids.
This has led to a large body of work on how to make predictions
accessible to decision makers, whether through models 
that are {\em  interpretable} \cite{lakkaraju2016interpretable}, 
or through {\em explainable machine learning},
%which has seen recent growth in popularity,
in which machine outputs (and so human inputs) are assumed to be predictions and are augmented with explanations \cite{ribeiro2016should,lundberg2017unified}.
%, likely due in part to the appeal of general, model- and task-agnostic tools \cite{doshi2017towards, lipton2016mythos}.
%Typical approaches are to provide additional information about  model internals \cite{angelino2017learning, lakkaraju2016interpretable}, or {\em post hoc} relationships between inputs and outputs \cite{ribeiro2016should, smilkov2017smoothgrad, lei2016rationalizing}.
% Important as it is, 
We see two main %a number of
drawbacks to these approaches. 
% \dcp{need to also critique on the first, ``interpretable ML`` approach} \sophie{is it unreasonable to just extend the existing critique?}% when humans are  the final arbiters.
%1. machine as expert
First, 
setting the role of machines to  `predict, then explain'
%removes agency from human decision makers. %Furthermore, explanations are primarily intended to aid in knowing when to accept or reject %model predictions. Not only are they of unsubstantiated value in this role, this %organizational structure reduces
% serves to 
reduces humans to auditors of the `expert' machines \cite{lai2018human}. With loss of agency, people are reluctant to adopt predictions and even inclined to go against them
 %---even in situations where humans knowingly underperform machine predictions on average
\cite{bandura1989human,bandura2010self, yeomans2017making,dietvorst2016overcoming, yin2019understanding, green2019principles}.
% brehm1966theory,
% especially after seeing the algorithm make errors, 
%s it may not be possible to describe complex models in a concise manner, and for this reason, many implementations fail to generate user trust and understanding \cite{lage2019evaluation}.
This leads to a degradation in performance of the human-machine pipeline over time \cite{elmalech2015suboptimal, dietvorst2015algorithm, logg2017theory, stevenson2018algorithmic}.
%noti2014experimental,
%Furthermore, difficulties in describing complex models in a concise manner
%result in implementations that fail at generating user trust and understanding \cite{lage2019evaluation}.
%Essentially, in any system in which humans make the final decisions, even highly-accurate machine outputs are only useful if
%and when humans make appropriate use of them,
%cf.  the use of risk assessment tools in the context of sentencing \cite{stevenson2018algorithmic}.
%2. non-adaptive to human actions; trained before, not during
More importantly, these methods 
%are non-adaptive in that they do not account for 
cannot adapt to the ways in which
 predictions are used, and so are unable to adjust for systematic human errors 
%  or inconsistencies 
 or to make use of human capabilities.

Moving beyond predictions,
in this paper we advocate for broader forms of learnable advice and  capitalize  on a different
strength of machine learning:
the ability to learn useful \emph{representations}.
% But learning algorithms are also adept at learning \emph{representations},
% and in this paper we capitalize on this idea to support decision-making.
% As simple machine predictors are limited, so, too, are human arbiters.
Inspired by the success of  representation learning, in which deep neural networks learn data representations that enable `simple'
(i.e., linear) predictors to perform well \cite{bengio2013representation},
we leverage neural architectures to learn representations that best support human decision-makers
% and are trained to align with human cognitive strengths and weaknesses
\cite{kahneman2011thinking,miller1956magical}.
%
% who may be analogously constrained (e.g. by limited cognitive capabilities and 
% vulnerabilities to psychological biases
% Given that pattern recognition is the hallmark of machine learning,
% %and bias mitigation are problems at which computers excel,
% there is a clear opportunity for data-driven, human-computer collaboration.
% Consider a multi-layered neural network $\nn(x)$. % parameterized by $\repprm \in \Repprm$.
% To support human decision-makers, 
Consider a multi-layered neural network $\nn = f \circ \repnet$
composed of a high-dimensional representation mapping $\repnet$
and a predictor $f$. Our key proposal is to remove the predictor and instead plug
the \emph{human decision function} $h$ into the learning framework to obtain $h \circ \repnet$,
allowing us to optimize the representation mapping
% $\repprm$ 
to directly improve human performance. 

Our framework for optimizing $h \circ \repnet$,
which we refer to as `Mind Composed with Machine' (\mom)
contributes to work that seeks to
bridge  machine learning with  human-centric design~\cite{sutton2020overview,venkatesh2003user},
% davis1989user,
% and advocate for a broader perspective on using machine learning
% in support of decision-making.
%
and we make two key contributions in this regard.
%that differs from the current paradigm in two important ways.
%
First, rather than  machines that predict  or decide,
we  train models that learn how to \emph{reframe problems}
for a human decision-maker. We learn to map problem instances to representational objects such as plots, summaries, or avatars,
aiming to capture  problem structure and preserve user autonomy.
% For example, our framework can learn to map high-dimensional inputs into
% visual scatterplots, so that users can apply spatial reasoning to reach an appropriate conclusion.
%Prediction can be though of as reframing problems as scalars (e.g., predictive probabilities)
% \dcp{again, drop 1-2 cites if poss} 
This approach of ``advising through reframing" %Our approach
draws on work in the social sciences
% builds on a key observation, in the social sciences
that shows that the quality of human decisions depends  on how problems are presented
\cite{thompson1980margaret,cosmides1992cognitive,gigerenzer1995improve,kahneman2013prospect, brown2013framing}.
% cao2017statistically,
%As advice, prediction is but a narrow form of problem representation,
%Our goal here is to explore broad notions of problem reframing,
%and particularly those in which structural aspects of the problem are preserved.
%
Second, rather than optimizing for machine performance,
we \emph{directly optimize for  human performance}.
We learn representations of inputs
for which human decision-makers perform well
rather than those under which machines achieve high accuracy.
In this, we view our approach as taking a step towards
promoting machine learning as a tool for human-intelligence augmentation~\cite{licklider1960man,engelbart1962augmenting}.

The immediate difficulty in learning human-facing representations in \mom\ is that $h$ encodes how actual human decision-makers respond to representational advice
and so is not amenable to differentiation (we cannot ``backprop through $h$.")
To overcome this, we propose an iterative human-in-the-loop procedure
that alternates between
(i) learning a differentiable {\em  surrogate model} of human decision-making
% using human decisions 
at the current representation,
and (ii) training the machine model end-to-end 
using the current surrogate.
% to update the representation and improve decisions.
%(e.g., by changing the 2D projection)
For estimating the surrogate model we query actual humans for
their decisions given a current representation.
We demonstrate the \mom\ framework  on three distinct  tasks,
designed with two goals in mind:
to explore different forms of human-facing representations
and to highlight different benefits that come
from the framework.
%
%\sophie{do we need these descriptions here? an almost identical section precedes the experiments} 
The first experiment focuses on classifying {\em point clouds} %in which the appropriate representation class and parameterization are known,
in a controlled environment.
Here we show how the \mom\ framework can learn
scatter-plot representations that allow for high human accuracy without explicitly presenting machine-generated predictions (or decisions).
% Here, inputs are high-dimensional point clouds, represented to users through 2D scatterplots,
% and our framework learns through human feedback a 2D projection leading to optimal human performance.
%
The second experiment  considers 
loan approvals and adopts {\em facial avatars } as the form of representational advice.
Here we  demonstrate that the framework can be applied at scale
(we train using $\sim$ 5,000 queries to Amazon mTurk)
% \nir{sophie is this number correct?}
% \sophie{it's reasonably close}
and also explore 
what representations learn to encode
% the kinds of representations that are learned
% in this unique form of representational advice
and how these representations are used to support human decision-making.
% ,
% \niradd{and with large-scale human-in-the-loop training using Amazon
% Mechanical Turk.
%Here, we take care to ensure users understand that the avatar represents an algorithmic assistant and not a loan applicant.
%and shows that representations can learn to present feature information that improves human performance even when the user population is untrained and varies from round to round. \dcp{again, not hitting the spot because it is not specific enough to what we do} 
% As we later discuss, there are limitations to using
% Turkers for extending human capabilities, 
The third experiment is designed to demonstrate the capacity of our framework
to support decision-making in ways that outperform either human or machine alone. Here we use a simulated environment to show how \mom\ can learn a representation that enables a
human decision-maker to  incorporate \emph{side-information} (consider e.g. a hospital setting, in which doctors have the option to run additional tests or query the patient for information not included in the machine model), 
even when this information
is known only to the user. 
% \dcp{provide the health risk / ER context of this setting }
% }
% \dcp{need to justify simulation here. best to get ahead of this}
%we show that optimizing for human performance results in better models than those trained for %prediction.
%\blue{The second task uses a synthetic human model to demonstrate that features learned by our framework are better aligned with human causal models than those generated by a popular black-box explanation tool.}
%
% \dcp{do we need li and jordan?} 

% ,li_2018,jordan_2018}.
%\nir{are we setting expectations too high? will reviewers be disappointed when they read the experiments section?}

{\bf On the use of facial avatars:} 
% \dcp{reordered this section; i think it's stronger like this} 
In our study on loan approval we convey advice through a facial avatar that represents an algorithmic assistant.
We take care to ensure that users understand this, and understand
that the avatar does {\em not} represent a loan applicant. We also restrict
the avatar to carefully chosen variations on the image of a single actor.
We are interested to experiment with facial avatars as representations because facial avatars  are high dimensional,  abstract (i.e., not an object that is in the domain studied), and naturally accessible to people. We are aware of the legitimate concerns regarding the use of faces in AI systems and the potential for  discrimination~\cite{west2019}
and any use of facial representations in  consequential decision settings must be done with similar 
care.

\section{Related Work} \label{sec:related}

% \todo{REVISE, look for recent work to cite}\\
% \todo{new citatoins?  MGR PNAS 2019, lipton? others?}\\
% \todo{merge with related section from Neurips full paper submission}\\
% \todo{...}

% \subsection{Modeling human factors and machine arbiters.} 
\subsection{Modeling Human Factors} 
Recent studies have shown that the connections between trust, accuracy, and explainability can be complex and nuanced. Human users tend to use algorithmic recommendations less frequently than would be beneficial ~\cite{green2019disparate, lai2018human}, and user trust (as measured by agreement with algorithmic recommendation) does not increase proportionately to model accuracy \cite{yin2019understanding}.
% \dcp{place?} trust in a model even 
% when model accuracy is superior to human accuracy 
Increasing model interpretability may not increase trust (as measured by agreement with the model), and may decrease users' ability to identify model errors \cite{ poursabzi2018manipulating}. Further, even when explanations increase acceptance of model recommendations, they do not increase self-reported user trust or willingness to use the model in the future \cite{cramer2008effects}. In fact, explanations increase acceptance of model recommendations even when they are nonsensical \cite{lai2019human} or support incorrect predictions \cite{bansal2020does}.
%dcp cut space For these reasons, it may seem desirable to avoid using the human's decisions at all when it is avoidable.
% Expertise has been identified as a potentially confounding factor \cite{logg2017theory}
% when human experts wrongly believe they are better than machines,
% or when they cannot incorporate domain-specific knowledge within the data-driven model estimate.
At the same time, understanding human interactions with machine learning systems is crucial; for example, whether 
or not 
users retain agency has  been shown to affect users' acceptance of model predictions \cite{dietvorst2016overcoming}, providing support for
our approach. % that we take with \mom. 
Recent work acknowledges that human decision processes must be considered when developing decision support technology~\cite{lai2020harnessing, bansal2019beyond}, and work in cognitive science has shown settings in which accurate models of human decision-making can  be developed~\cite{bourgin2019cognitive}.

\subsection{Humans in the Loop}
%\dcp{following is generally good, although we might drop the final sentence in the para to be safe. need to fix refs}
Despite much recent interest in training with ``humans in the loop,"
experimentation in this setting remains an exceptionally challenging task.
The field of interactive machine learning has successfully used human queries to improve machine performance in tasks where human preferences determine the gold standard \cite{amershi2014power}, but human-in-the-loop training has been less productive in adapting predictive machines to better accommodate human decision-makers. 
% \dcp{ambigious: "a machine that outperforms" or failing to achieve a combination that outperforms humans alone?}.
In the field of interpretable machine learning, optimization for human usage generally relies on
% training 
% to optimize 
proxy metrics of human interpretability in combination with machine accuracy~\cite{lage2019human}, with people only used to evaluate performance at test time. 
% \dcp{`primarily`? is there something we need to mention in more detail?}
 A few exceptions have allowed human feedback to guide model selection among similarly-accurate machine-optimized models \cite{ross2017right, lage2018human}, incorporating human preferences. 
%  \dcp{ambiguous... break ties between models that are all equally accurate, or something else?} 
 In regard to using human responses as part of a feedback loop to a learning system, 
%  , which we discuss further in relation to human in the loop training.
%  We are not aware of any work in this space which models humans in the machine training procedure.
%Training with humans in the loop is an exceptionally challenging task,
%one that becomes even more pronounced in our setting:
%training complex machine models, driven solely by human actions, and using a novel interface between man and machine.
%Our experiments required substantial work to design, set up, and test,
%% (especially given that each experiment includes a very different instantiation of the framework)
%and are the product of much deliberation and careful balancing of many considerations.
%%
%And while we agree that experimental design should strive for high degrees of realism, diversity, human participation, and societal significance,
%we would like to place our efforts in the context of the current state of practice:
% As mentioned above, \citet{lage2018human} is the only interpretability method we know of to directly incorporate human responses into training. 
% Much of the interpretability literature uses humans for evaluation, but does not train with humans in the loop---the only exception we are aware of is \red{[Lage 2018]},
% which uses a much simpler setting 
 we are only aware of \citet{lage2018human}, and the authors actually abandoned attempts to train with mTurkers.
%  working instead with a group of machine learning graduate students and postdocs.
% \dcp{consider dropping next sentence. too risky and not needed?}  We believe our work is the first to use human responses during training to optimize human performance, evaluating at scale with real users.
% Other work on human-machine cooperation  proposes methods for training with humans, but either demonstrates results through fully synthetic experiments using simulated human responses \cite{madras2018predict}, 
% or makes use of previously collected human datasets \cite{wilder2020learning}. (this is not applicable for our setting, as we need to query arbitrary points in representation space).
% Our framework directly optimizes for human performance
% by querying humans during training for their decisions on arbitrary points in representation space.
%dcp cut for space 
%Our work
% We are not aware of any prior work that
%optimizes representations directly for human performance
% nor of any papers that successfully use %the highly variable responses of mTurkers for large-scale human-in-the-loop training.
%using human-in-the-loop training.

%\dcp{why `interpretability` and not `explainability`?} 
\subsection{Collaboration with Machine Arbiters}
% \todo{wilder, mozannah sontag}
A related field considers learning when a machine learning system  should defer to a human user instead of making a prediction. This setting, unlike ours, allows the machine to bypass a human decision-maker ~\cite{madras2018predict, mozannar2020consistent, wilder2020learning}. 
% \dcp{unclear what `human arbitration` means} 
In this setting, human accuracy is considered to be fixed and independent of the machine learning system, and in evaluation human decisions are either fully simulated or based on  previously gathered datasets.

\section{Method} \label{sec:method}
% 

%We begin by introducing a formal framework for algorithmic decision support,
%through which we highlight the limitations of current approaches,
%and present our proposed solution.

\begin{figure*}[t]
	\begin{center}
		\includegraphics[width=0.95\textwidth]{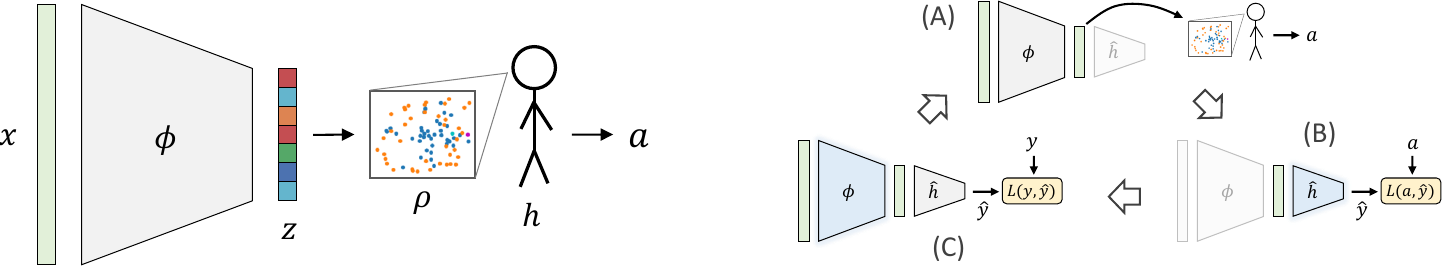}
		\caption{		\textbf{Left}: The  \mom\ framework.
			The neural network learns a mapping $\repnet$ from inputs $x$ to representations $z$,
			such that when $z$ is visualized through
			$\vizmap$, representations elicit good human decisions.
			\textbf{Right}: Training alternates between
			(A) querying users for decisions on the current representations,
			(B) using these to train a human surrogate network $\hhat$,
			and (C) re-training representations.
			\label{fig:illustration} }
	\end{center}
\end{figure*}

In a typical setting, a decision-making user is given an
{\em instance} $x \in \X$.
For clarity, consider $\X=\R^d$.
Given $x$, the user must decide on an {\em action} $\des \in \Des$.
For example, if $x$ are details of a loan application, 
then users can choose $\des \in \{\texttt{approve},\texttt{deny} \}$.
Each instance is also
associated with a ground-truth \emph{outcome} $y \in \Y$, 
% \dcpadd{
so that $(x,y)$ is sampled from an unknown distribution $D$.
% }.
We assume that users seek to choose actions that
minimize an incurred {\em loss} $\loss(y,\des)$, with $\loss$ also known to the system designer; e.g., for loans, $y$ denotes
whether a loan will be repaid.
% or not a person will repay the loan.
%
 We consider
%  \dcpadd{
 the general class of
%  }
\emph{prediction policy problems}~\cite{kleinberg2015prediction},
where the loss function is known and the difficulty in decision-making is governed by 
how well $y$ can be predicted.

We  denote by $h$ the {\em human mapping} from  inputs to decisions or actions.
For example, 
$\des=h(x)$ denotes a decision based on raw instances $x$.
Other sources of input  such as {\em explanations} $e$ or
representations can be considered;
e.g., $\des=h(x,\yhat,e)$ denotes a decision based on $x$ together with prediction $\yhat$ and explanation $e$. 
% \dcp{is it worth a FN to explain that this is different from how we described `explainable ML` earlier? there, the explanation went along with a prediction from the machine, not $x$} 
We allow $h$ to be either deterministic or randomized, and conceptualize 
$h$ as either representing  a particular target user or a stable distribution over different kinds of users. We assume  the mapping $h$ is fixed (if there is adaptation to a representation, then $h$ can be thought of as the end-point of this adaptation).
%additional side information $s$ via $h(x,s)$,
%a machine-generated prediction $\yhat$ via $h(x,\yhat)$
%or explanation $e$ via $h(x,e)$).
% \todo{
% -	say we assume h is fixed (so no human learning)\\
% - clarify - h as model of single human vs crowd average
% }

%Our goal as designers of a decision-support system is to aid users in choosing good actions.
Crucially, we also allow machines to present users with machine-generated \emph{advice} $\advice(x)$,
with human actions denoted as $\des=h(\advice(x))$. Users may additionally have access to {\em side information} $s$ that is unavailable to the machine, in which case user actions are $\des=h(\advice(x), s)$.\footnote{This notion of machine-generated advice  generalizes both explanations
(as $\advice = (x,\yhat,e)$, where $e$ is the explanation)
and deferrals (as $\advice = (x,\ybar)$, where $\ybar \in \{0,1,\text{defer}\}$, with a human model that always accepts $\{0,1\}$)~\cite{madras2018predict}. \sophie{do we want to separate ourselves from deferral?}}
%Both approaches, however, remain strongly tied to a predictive form of advice.
%whereas we advocate for considering forms of advice that are more flexible, expressive,
%and human-aligned.
%
% \dcp{introduce side information here or earlier}
%
%
%
Advice $\advice(x)$ allows for  a {\em human-centric representation} of the input, 
and we seek to \emph{learn} a mapping $\advice$ 
from inputs to representations under which humans will make good decisions.
%The main question we explore is what makes for good algorithmic advice $\advice$.
The benchmark for evaluation  is the expected loss of human actions
given this advice:
\begin{equation}
\label{eq:overall_objective}
\expect{D}{\loss(y,\des)}, \qquad \mbox{for}\quad   \des = h(\advice(x)).
\end{equation}

%\dcp{limitation of this set-up is that it does not make explicit that human may have additional input, say $x'$}

\subsection{Predictive Advice}
% \dcp{all of this goes through fine without requiring the machine target to be in $\Y$. crucial thing is that it relies on human being identity function wrt machine output}
%
%Our goal in this paper is to explore what constitutes good algorithmic advice.
A standard approach  provides human users with machine-generated predictions, $\yhat = f(x)$, 
%\sophie{and assume a known mapping from $\yhat \rightarrow \des \in \Des$} \footnote{note that when $\Des = \Y$, predictions are themselves action recommendations.}, 
%dcp let's drop any consideration of how predictions go to actions until as needed later 
where $f$ is optimized for predictive accuracy
and there is a straightforward mapping from predictions to prescribed actions $\hat{y} \rightarrow \hat{y}_{a}$
% \dcpadd{
% and the predictions themselves correspond easily to action recommendations 
(e.g., for some known threshold, `probability of returning loan' corresponds to `approve loan'). 
% \dcp{is it confusing that we suggest 0/1 predictions here, not `prob will return loan`?}
% }
This is a special case of our framework where advice $\advice=(x,\yhat)$,
% \dcpadd{and where it is  assumed that the user model is action
and the user is modeled as $a=\yhat_{a}=h(x,\yhat)$.
% }.
\sophie{check this}\dcp{comment that $x$ plays no role and this also models just providing the prediction? in fact, is it more standard for this to just pass along $\hat{y}$ and not also $x$? } The predictive model is
trained to minimize:
\begin{equation}
\label{eq:machine_objective}
\min\nolimits_f \expect{D}{\loss(y,\yhat_{a})}, \qquad \mbox{for}\quad  \yhat = f(x).
\end{equation}

% \dcp{can simplify following}
% While the objective in \eqref{eq:machine_objective} aligns with that in \eqref{eq:overall_objective} (i.e., both aim to lower $\loss$),
% learning with \eqref{eq:machine_objective} can affect overall performance only indirectly,
% as it hinges on the way in which users incorporate predictions into their decisions
% via $\des=h(x,\yhat)$.
% Hence, predictions are useful only to the extent that the human decision-maker will follow them,
% and the success of optimizing \eqref{eq:machine_objective} as a proxy for
% \eqref{eq:overall_objective} effectively relies on excessive (and often unrealistic)
% assumptions on the compliance of humans. %$h$.

%Optimizing \eqref{eq:machine_objective} is 
%This is equivalent to optimizing \eqref{eq:overall_objective} for a specific user model, $\yhat=h(x,\yhat)$, and

In this approach, predictions $f(x)$ are useful only to the extent that they are followed.  Moreover, 
predictions provide only a scalar summary of the information
in $x$, and limit the degree to which users can exercise their cognitive
and decision-making capabilities; 
% \dcpadd{
e.g.,  in the context of side information.

\subsection{Representational Advice}
%Both approaches, however, remain strongly tied to a predictive form of advice.
%whereas we advocate for considering forms of advice that are more flexible, expressive,
%and human-aligned.
%%
%We propose a fundamentally different approach for the use of machine learning to promote good decisions in settings with human users.
%, that differs in two significant ways.
%First,
In \mom, we  allow advice $\advice$ to map inputs into representations that are designed to usefully convey information to a human decision-maker (e.g., a scatterplot,
a compact linear model, or an avatar).
% \dcpadd{
Given a {\em representation class}  $\Advice$ we  seek a  mapping $\advice\in\Advice$  that minimizes expected loss
$\min_{\advice \in \Advice} \expect{D}{\loss(y,h(\advice(x)))}$.
%\end{equation}
%
%where is a class of parameterized representation mappings.
%from which we aim to learn the optimal $\advice \in \Advice$.
%In practice, queries to humans $h$ are limited in number and precision (we return to these issues in the algorithm), but assuming for the moment that we have oracle access to $h$ and its gradients, to learn optimal $\advice$ 
%
% \dcp{explain why}
%
With a {\em training set} $\smplst=\{(x_i,y_i)\}_{i=1}^m$
sampled from distribution $D$, and with knowledge of the human mapping $h$, we would seek $\advice$ to minimize the
\emph{empirical loss}:
\beq
\label{eq:objective}
\min_{\advice \in \Advice} \sum_{i=1}^m \loss(y_i, \des_i), %+ \lambda \reg(\advice),
\qquad \mbox{for}\quad 
\des_i = h(\advice(x_i)),
\eeq
 possibly under some form of regularization (more details below). \sophie{check} \dcp{explain , e.g. in FN, what role regularization will play}
% \dcp{will be confusing, because we don't talk directly here to also having training data on what humans do with advice. extend discussion here?}
%$\reg$ is a regularization term,
% % that can be task-specific and data-dependent
%  and $\lambda$ is the regularization parameter.
%
%
%\subsubsection{Human-aligned representations.} 
%What makes for good representations?
%
Here, $\Advice$ needs to  be rich enough to contain
 flexible mappings from inputs to representations while also generating objects that are accessible to humans.
To achieve this, we decompose  algorithmic 
advice $\advice(x) = \vizmap(\repnet_\repprm(x))$ into two components: %, where 

$\bullet$ $\repnet_\repprm:\R^d \rightarrow \R^k$ is a  parameterized \emph{embedding model} 
with learnable parameters $\repprm \in \Repprm$, that maps  inputs into vector representations
$z = \repnet_\repprm(x) \in \R^k$ for some $k>1$, and  

$\bullet$ $\vizmap : \R^k \rightarrow \vizobjs$ is a   \emph{visualization component} 
that maps each $z$ into a  visual object $\viz = \vizmap(z) \in \vizobjs$
(e.g., a scatterplot, a facial avatar).
\smallskip

This decomposition is useful because for a given application of \mom\ we 
can now fix the visualization component $\vizmap$, and  seek to learn  
the embedding  component $\repnet_\repprm$. 
This process of learning a suitable embedding through feedback from human users, is 
what we mean by ``learning representations by humans [from feedback], for humans." 
% \dcp{right?}
\dcp{suggest also  giving quick examples here of $\vizmap$ to fix ideas for the reader about what we do/don't do. i know some readers get dissuaded at this point}
\sophie{these are given above - is it too easy to miss? should move to main text?}
%which we will implement using deep neural networks.\footnote{
% \dcp{propose to cut next sentence}We believe this to be a reasonable simplification, as we are not aware of a method to parameterize over a wide space of visualizations.
%dcp cut , and for a given visualization method, it is without loss of generality for parameters within the visualization to be pushed outside and used as inputs.}
%\dcp{discuss here why reasonable to start with assuming vizmap fixed}
Henceforth, it is convenient to fold the visualization component $\vizmap$ into the human mapping $h$,
and write $h(z)$  to mean $h(\vizmap(z))$, for embedding $z=\repnet_\repprm(x)$.
%$h^{(\vizmap)}(z) = h(\vizmap(z))$.
The training problem~\eqref{eq:objective} becomes:
\beq
\label{eq:objective_v2}
\min_{\repprm \in \Repprm} \sum_{i=1}^m \loss(y_i, \des_i), \quad \mbox{for}\
\des_i = h(\repnet_\repprm(x_i)),
\eeq
%\dcp{explain $R$}
%
again, perhaps with some regularization.
By solving \eqref{eq:objective_v2},
we learn representations that promote good decisions by the human user.
%We refer to this framework as {\em Man Composed with Machine} (\mom), illustrated in 
See Figure~\ref{fig:illustration} (left). 

\paragraph{Regularization.} Regularization may play a number of different roles: as with typical L2 regularization, it may be used to reduce overfitting of the representation network, encouraging representations that generalize better to new data points. It may also be used to encourage some desired property such as sparsity, which may be beneficial for many visualizations, given the limited ability of human subjects to process many variables simultaneously. Regularization can also be used in our framework to encode domain knowledge regarding desired properties of representations, for example when the ideal representation has a known mathematical property. We utilize this form of regularization in Experiments 1 and 2.

\paragraph{Choosing Appropriate Visualizations.} 

Determining the form of representational advice that best-serves expert decision-makers in any concrete task will likely require in-depth domain knowledge and should be done with care. The characterization of varying visualizations' effects on decision-making is sufficiently elaborate as to warrant its own field of study \cite{lurie2007visual}, and thus we focus here on learning to adapt a particular choice of representation from within a set of ``approved" representational forms.

%\subsection*{Method}
%Deep neural networks can be conceptualized as powerful tools for learning
%vector representations under which simple predictors (i.e., linear) perform well %\cite{bengio2013representation}.
%In this work,
\subsection{Training Procedure, and Human Proxy}
%dcp cut for space
%Whereas representation learning has come to mean finding representations of an input that allows machine predictors to perform well \cite{bengio2013representation}, We   apply representation learning to a setting where people and not machines consume the representations. 

%where a good representation is one that alleviates the difficulty of working directly with $x$.

%dcp cut for space
%Consider a multi-layered neural network $\nn(x)$. % parameterized by $\repprm \in \Repprm$.
%Splitting the network at some layer partitions it into 
%a parameterized representation mapping $\repnet_\repprm:\R^d \rightarrow \R^k$
%and an action selection function $\prednet : \R^k \rightarrow \Des$, 
%  with  $\nn(x)=\prednet(\repnet_\repprm(x))$. 
%To support human decision makers, we remove $\prednet$ and instead plug in
%the human decision function $h$,
%therefore leveraging the optimization of $\repprm$ to directly improve human performance.

%\subsection*{Optimization}
%The main difficulty is that 

We  adopt a neural network to model the 
parameterized embedding $\repnet_\repprm(x)$, and thus advice $\advice$.
The main difficulty in optimizing \eqref{eq:objective_v2} is that
human actions $\{\des_i \}_{i=1}^m$ depend on $\repnet_\repprm(x)$
via an unknown  $h$  and yet gradients of $\repprm$ must pass through $h$.
%
%For this reason, we also  need to learn a {\em proxy} for
%the human mapping $h$.
%, \dcpadd{as $h$ is unknown, and even if it was known, may not be amenable to optimization.}
%
%
%
%
To handle this, we %also need to learn a
make use of a
{\em differentiable surrogate for $h$}, denoted 
$\hhat_\hhatprm : \R^k \rightarrow \Advice$ with parameters $\hhatprm \in \Hhatprm$.  We learn this surrogate,
referring to it as ``h-hat." 
%For optimizing model components (i.e., $\repprm$ and $\hhatprm$),

The \mom\ {\em human-in-the-loop training procedure} alternates between  two steps:
\begin{enumerate}
\item Use the current $\repprm$ to gather samples of  human decisions $\des = h(z)$ on inputs $z=\repnet_\repprm(x)$ and fit $\hhat_\hhatprm$.
\item Find $\repprm$  to optimize the performance of
$\hhat_\hhatprm \circ \repnet_\repprm$ 
for the current $\hhatprm$, as in \eqref{eq:objective_v2}. 
%\dcp{check: `find .. to optimize` ok? or is it more about following gradients?}
\end{enumerate}
\smallskip

% \input{./chapter1/pseudocode}

% \dcp{definitely by here need to have mentioned that training data is augmented with human decisions}

Figure~\ref{fig:illustration} (right) illustrates this process;
% and pseudocode is given in Algorithm \ref{algo:pingpong}.
for pseudocode see Appendix~\ref{sec:algo}).
Since $\hhat$ is trained to be accurate for the current embedding distribution rather than globally, $\hhat$ is unlikely to exactly match $h$. \sophie{updated}\dcp{explain what `local` means here; ambiguous, e.g. we don't mean ` local minima` in sense of non-convex problems!}
However, for learning to improve, 
% This is not a problem,
it suffices for $\hhat$ to  induce parameter gradients  that improve loss  (see Figure \ref{fig:hhat}
in the Appendix).
Still, $\hhat$ must be periodically retrained because
% as the training of $\repnet$ proceeds,
as  parameters $\repprm$ change, so does the induced distribution of %vector representations
representations $z$ (and  $\hhat_\hhatprm$ may become less accurate).
% and $\hhat_\hhatprm$ may no longer be representative of human responses. % on this new distribution.
% \dcp{do we have any analysis anywhere of the ability to be robust to inaccuracy in hhat? if so, point to here in a footnote}
%\todo{ref appendix on robustness to hhat not perfect vs h}
% Further details about this robustness to inaccurate h-hats, as well as  pseudocode, can be found in Appendix~\ref{sec:algo}.

% Note that for the performance to improve,
% it suffices for $\hhat$ to induce gradients of the loss
% %$\nabla_\repprm \hhat(\repnet_\repprm(x))$
% that approximate those of $h$.
% That is, $\hhat$ need not exactly match $h$ and . \dcp{would be really good to illustrate this point that $\hhat$ we learn are not accurate, but still help, experimentally}

%\nir{add section on what are good viz reps?}
%\dcp{yes, good idea}

\paragraph{Initialization of $\repprm$.}

In some applications, it may be useful to initialize $\repnet$ using a machine-only model with architecture equal to $\hhat(\repnet)$. In applications in which the human must attend to the same features as the machine model, this can help to focus $\repnet$ on those features and minimize exploration of representations that do not contain decision-relevant information. 
This can be particularly useful when the representation lies within the domain of the data (e.g. plots, subsets). 
% However, in domains in which it is possible for the machine-only setup to produce a high-accuracy model which relies on features inaccessible to human users (consider, for example, adversarial features in image recognition) this may focus $\repnet$ too narrowly  too early, making it more difficult to discover representations useful to humans. We note that the machine-only model with architecture equal to $\hhat(\repnet)$ may be otherwise useful in model selection: the architecture of $\repnet$ may be verified in a machine-only setting to be sufficiently flexible to achieve a desired representation distribution, and the architecture of $\hhat$ may be similarly verified to be capable of mapping a set of representation distributions to a set of binary answers with high accuracy.

When a desired initial distribution of representations is known, $\repnet$ can be positioned as the generator of a Wasserstein GAN \cite{arjovsky2017wasserstein}. In this case, the labels are not used at all, and thus the initial mapping is used only to achieve a certain coverage over the representation space and not expected to encode feature information from a machine-only model.

\subsection{Handling Side Information}

One way humans could surpass machines is through access to \emph{side information} $\side$ that is informative of outcome $y$ yet unknown to the machine.
The  \mom\ framework can  be extended to learn
% Under certain conditions, when the human has access to side information $s$ that carries signal about $y$, our method can adapt to find the
a representation $\advice(x)$ that is optimal 
conditioned on the existence of $\side$,
% this side information, 
despite the machine having no access to $\side$.
At test time, the human has access to $\side$, and so action $\des = h(\repnet(x),s)$.
% to approximate $h$ for a given $\repprm$.
The observation is that the ground-truth outcome $y$, which is available during training, conveys information about $s$:
if $\side$ is informative of $y$, then there exist $x$ for which the outcome $y$ varies with $\side$. Thus $(x,y)$ is jointly informative of $\side$: for such $x$, knowing $y$ and modeling the mechanism $y = g_{x}(s)$ by which $\side$ affects $y$ for a given $x$ would allow reverse-engineering the value of $\side$ as $g_{x}^{-1}(y)$.
% \dcp{explain this}
Although $\side$ cannot generally be exactly reconstructed without supervision on $\side$ (e.g. due to inexact modeling or non-invertibility of $g_{x}$),
in some cases $(x,y)$ can be used to make useful inference about $\side$. Intuitively, note that for a given $x$, multiple $y \in \{y_1 \hdots y_k\}$ values correspond to multiple $s$ values. If $h$ varies with $s$, without access to $s$ or $y$, the best $\hhat(x)$ we can learn is $\expect{s\sim S}{h(x,s)}$. With varied $y_i$ which correspond to different values of $s$, we can learn $\hhat(x, y_i) = \expect{s\sim S|y=y_i}{h(x,s)}$ for each $y_i$, which allow $\hhat$ to incorporate information about $s$.
\dcp{if possible, try to make this a bit more intuitive?}

\section{Experimental Results} \label{sec:experiments}

% In this section, 
We report the results of three distinct  experiments.
%, each
%designed to demonstrate different aspects of the \mom\ framework.
Our intent  is to demonstrate the breadth of the framework’s potential,
and the experiments we present vary in the
% decision 
task,  the form of 
% representational 
advice,
their complexity and scale,
and the degree of human involvement
(one experiment is simulated,
another uses thousands of mTurk queries).
%
%dcp cut for space The results demonstrate the generality of \mom\
%in the type of advice it allows and the decision tasks  it supports.
%We next describe each of these experiments,
We defer some of the experimental details  to the Appendix. 

\paragraph{Model Selection}
Experimenting with humans in-the-loop is expensive and time-consuming, making standard practices for model selection such as cross-validation difficult to carry out. This necessitates committing to a certain model architecture at an early stage and after only minimal trail-and-error. In our experiments, we rely on testing architectures in a machine-only setting with various input and output distributions to ensure sufficient flexibility to reproduce a variety of potential mappings, as well as limited human testing with responses from the authors. Our model choices produced favorable results with minimal tuning. We believe this suggests some useful robustness of the approach to model selection choices, but future work would be beneficial to better understand sensitivity to model selection.

% \nir{this was cut for space - should we bring it back in?} \dcp{prob ok to cut, and we'll need the space. we can provide any hand-holding as we get into each section}
% \blue{
% %\if 0
% In the first experiment,
% we focus on learning low-dimensional visualizations of high-dimensional data,
% showing that the \mom\ framework finds increasingly useful visualizations.
% %by utilizing human performance as feedback.
% %learn projections from high-dimensional problem instances to low-dimensional visualizations, and demonstrate how our system finds 
% In the second experiment,
% we focus on feature selection,
% and demonstrate through a simulation how \mom\ can learn an optimal linear model %subset of features
% in a setting where side information is available to users.
% %into their decisions.
% %we focus on text summarization, and compare the usefulness of our human-tailored summaries to those generated by ``explainable'' methods.
% In the third and largest experiment,
% we focus on loan approval, and convey algorithmic advice through facial avatars that are 
% trained to trigger gut-feelings that lead to good decisions. 
% We use real loan data, and train with decisions gathered from 
% subjects on Amazon Mechanical Turk. 
% We show that \mom\ can be effectively applied to a real decision context and with diverse users.
% %and to explore what representations may learn in this context.
% %
% %\fi
% }

% \dcp{note IRB here before moving on}

\subsection{Decision-compatible Scatterplots}

In the first experiment,
we focus on learning useful, low-dimensional representations of high-dimensional data, in the form of scatterplots. %in the present context.
%,
%showing that the \mom\ framework finds increasingly useful visualizations. \dcp{is it confusing to say `vis` and not `representation`? after all, the (method of) visualization $\rho$ is fixed}
%
%dcp cut for space 
To make high-dimensional data more accessible to users,
it is common practice to project into a low-dimensional embedded space and reason based on a visualization, for example a scatter plot or histogram.
%Visualizing high-dimensional data often involves a projection into a low-dimensional embedded space, where the data is then plotted. 
The choice of how to project high-dimensional data into a lower-dimensional space  is  consequential to decision-making  \cite{kiselev2019challenges},
and yet  
standard dimensionality-reduction methods  
optimize statistical criteria (e.g., maximizing directional variation in PCA) 
rather than optimizing for 
success in user interpretation.
The \mom\ framework learns projections that, once visualized,
directly support good decisions.
%directly from human feedback. 

We consider a  setting where the goal is
to correctly classify  objects in $p$-dimensional space, $p>2$.
Each $x$ is a $p$-dimensional point cloud
consisting of $m=40$ points in $\R^p$ (so $x \in \R^{40p}$).
Point clouds are constructed such that,
when orthogonally projected onto a particular linear 2D subspace of $\R^p$, 
denoted $V$,
they form the shape of either an `X' or an `O', this determining their true label $y$.
All directions orthogonal to $V$ contain similarly scaled random noise. In the experiment, we generate 1,000 examples of these point clouds in 3D.

% \dcp{we use embedding and projection. just use one in this section?}
%\todo{use projection, not embedding}
Subjects are presented with a series of scatterplots, which visualize the point clouds for a given 2D projection, and are asked to determine for each point  cloud its label (`X' or 'O').
% on the basis of its corresponding scatterplot.
Whereas a projection  onto $V$ produces a
useful representation, most others do not,
including those learned from PCA.
Our goal is to show 
that \mom\ can use human feedback to learn a projection ($\repnet$) that produces visually meaningful scatterplots ($\vizmap$), leading to good  decisions.

\paragraph{Model.}
Here, representation $\repnet$ plays the role of a dimensionality reduction mapping. We use $d=3$
and set $\repnet$ to be a 3x2 linear mapping
with parameters $\repprm$ as a 3x2 matrix.
This is augmented with an orthogonality penalty $\repnet^{T}\repnet - \mathbb{I}$ to encourage matrices which represent rotations.
For the human proxy model, we want to be able to roughly model the visual perception of subjects. 
For this, we use for $\hat{h}$ a small, single-layer 3x3 convolutional network, 
that takes as inputs a differentiable
6x6 histogram over the 2D projections. 

% For training, we use a fixed number of epochs (500 for $\hat{h}$ and 300 for $\repnet$) with base learning rates of .07 and .03, respectively, that increase with lower accuracy scores and decrease with each iteration. 
% We have found these parameters to work well in practice,
% but observed that results were not sensitive to their selection.
%\dcp{say a little here, eg., footnote, about how much experimentation went into this design of proxy or if it ``just worked"}
%\todo{flip order in prev sentence - intended to model visual perception, did this using conv net + histogram}
% \blue{Further details are described in Appendix \ref{sec:exp1}.}

%  For training, we use a fixed number of epochs (500 for $\hat{h}$ and 300 for $\repnet$) with base learning rates of .07 and .03, respectively, that increase with lower accuracy scores and decrease with each iteration. 
% We have found these parameters to work well in practice,
% but observed that results were not sensitive to their selection.
% % These parameters are not carefully chosen, as they were sufficient to make the task work in practice.
% The interface allows the number of rounds and examples to be determined by the user, but often 100\% accuracy can be achieved after about 5 rounds of 15 examples each.

\begin{figure}[t!]
	\includegraphics[width=0.47\columnwidth]{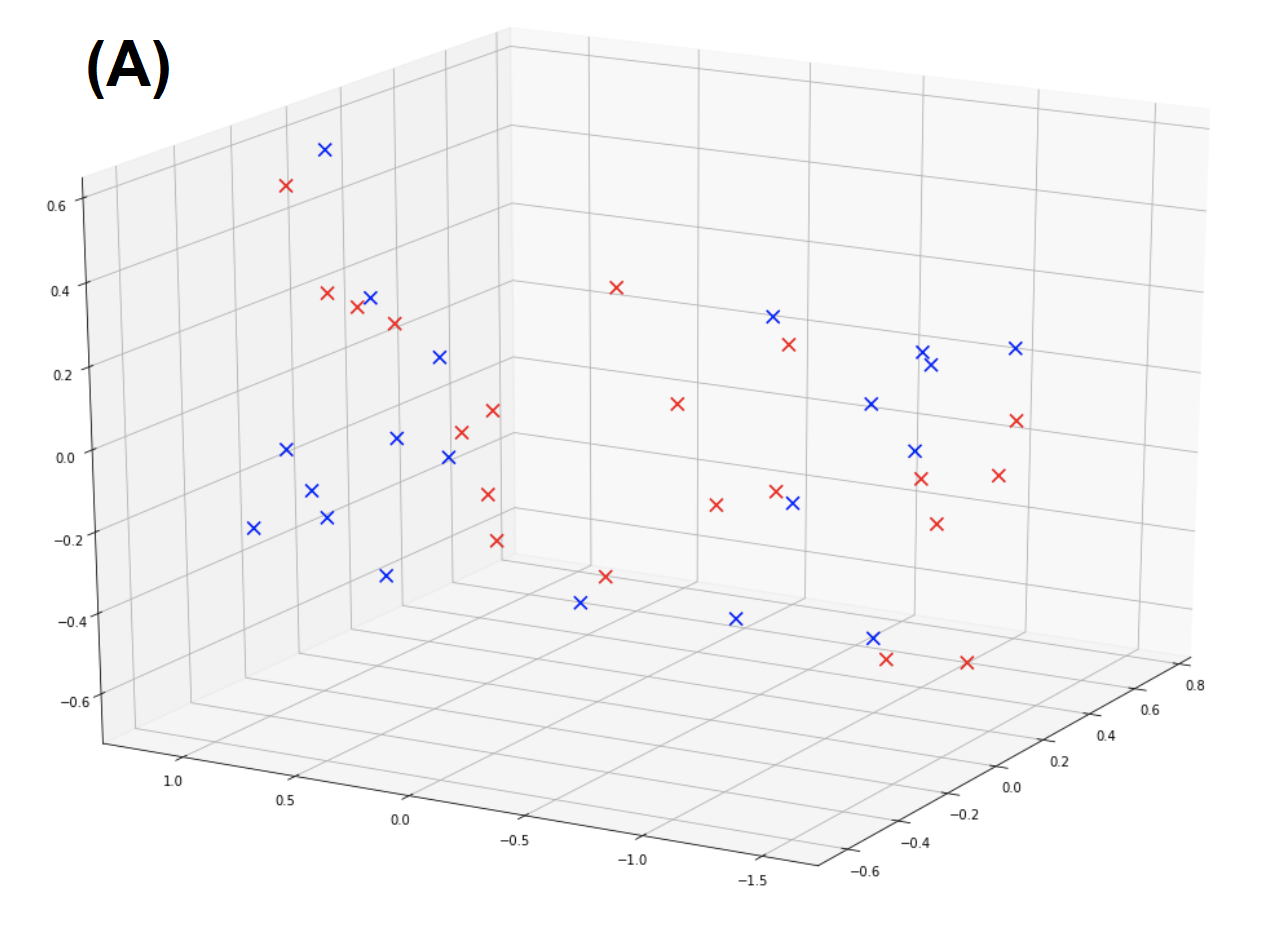}
	\includegraphics[width=0.47\columnwidth]{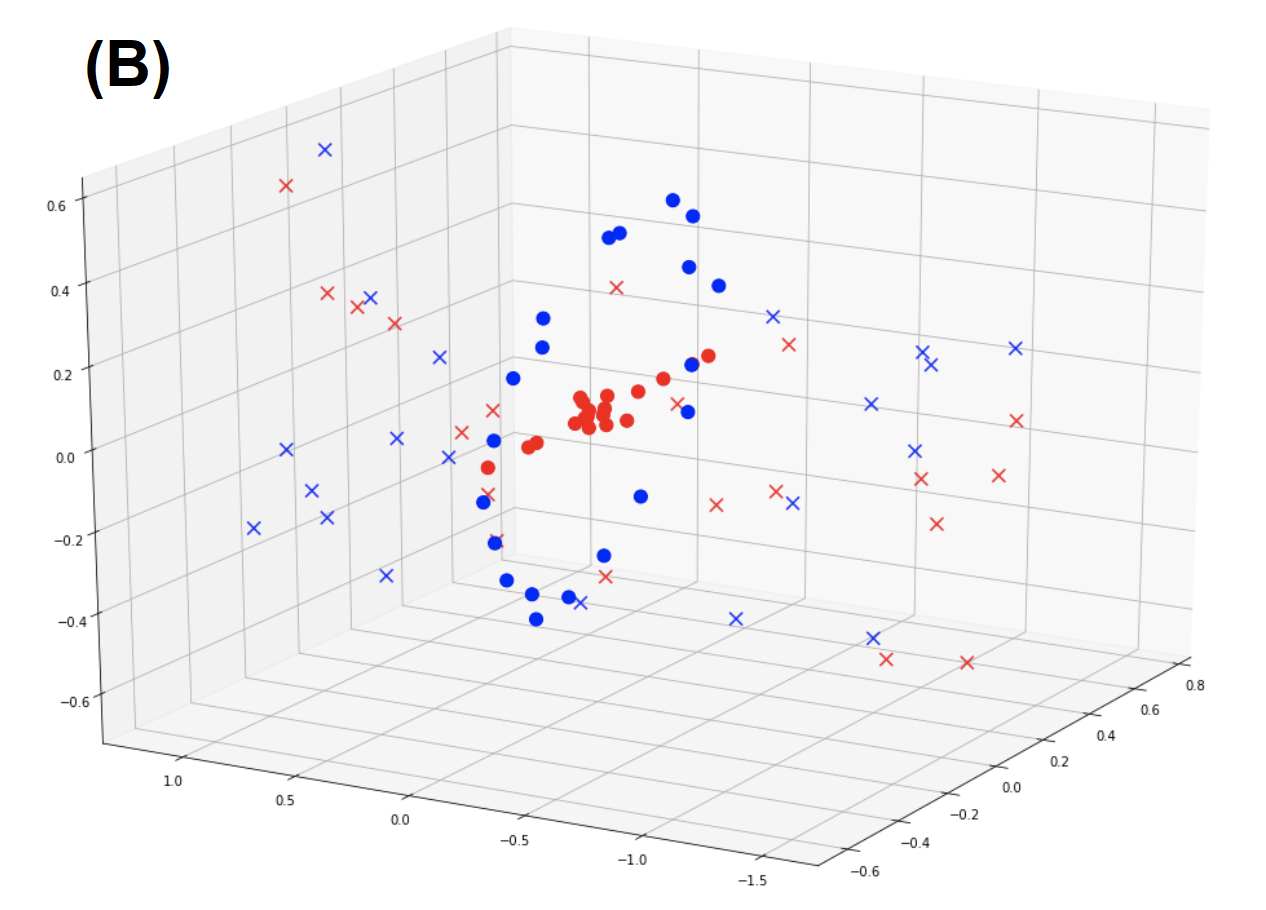}
	\includegraphics[trim=0 0 5cm 0, clip, width=1\columnwidth]{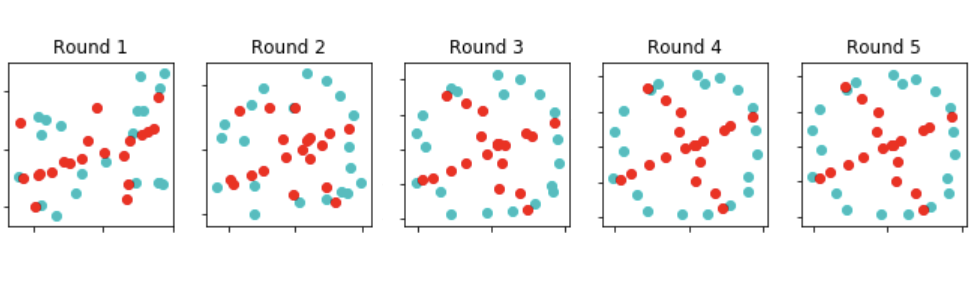}
	\caption{2D representations of point clouds.
		\textbf{(A)} Points in their original 3D representation give little visual indication of class (X or O).
		\textbf{(B)} Shapes become easily distinguishable when projected onto an appropriate subspace (shown in bold).
		\textbf{(Bottom)} Learned 2D representations after each training round (`X', `O' are overlaid).
		The initial 2D projection (round 1), on which a machine-classifier is fully accurate, 
% 		with the representation corresponding to a fully accurate machine-only model 
		is unintelligible to people.
		However, as training progresses, feedback  improves
		the  projection until the class becomes visually apparent (round 4),
		with very high human accuracy.}
	\label{fig:xo} 
\end{figure}

\paragraph{Results.}
%\todo{STUDY RESULTS}
We recruited 12 computer science students 
% \dcp{maybe remove the critique in related work about some authors who used people from their lab not turkers?} 
to test the  \mom\ framework.\footnote{All experiments are 
conducted subject to ethical review by the university's IRB.} Participants watched an instructional video and then completed a training and testing phase,
each having five rounds
(with intermittent model optimization)
of 15 queries to label plots as either `X' or `O'.
The results we provide refer to the testing phase.
Round 1 includes representations based on a random initialization
of model parameters and therefore serves as a baseline condition.
The results show that participants achieve an average accuracy of 68\% in round 1, but improve to an
average accuracy of 91\% in round 5,
a significant improvement of 23\% ($p<.01$, paired $t$-test)
with 75\% of participants achieving 100\% accuracy by round 5.
% \footnote{Improvement from training round 1 to testing round 1 is negligible, .03.}
%
Subjects are never given machine-generated predictions or feedback, and improvement from training round 1 to testing round 1 is negligible (3\%), suggesting that
progress is driven solely by the
% performance of subjects
% on the 
successful reframing of problem instances (not humans getting better at the task).

Figure \ref{fig:xo} demonstrates a typical example of a five-round sequential training progression.
Initially, representations produced by \mom\ are difficult to classify when $\repprm$ is initialized arbitrarily. (This is also true when $\repprm$ is initialized with a fully accurate machine-only model.)
As training progresses, feedback regarding subject perception
gradually rotates the projection,
revealing distinct class shapes. % (see Figure~\ref{fig:avatar_acc} top left). \dcp{pointing to avatar figure?}
%
% \dcp{sophie was nominated to rephrase next sentence!} 
Training progress is made as long as subject responses carry some machine-discernible signal regarding the subject's propensity to label a plot as `X' or `O'.
\mom\ utilizes these signals
to update the representations and improve human performance.

%=====================================================================================

 \begin{figure}[t]
	\centering
	\includegraphics[width=1\columnwidth]{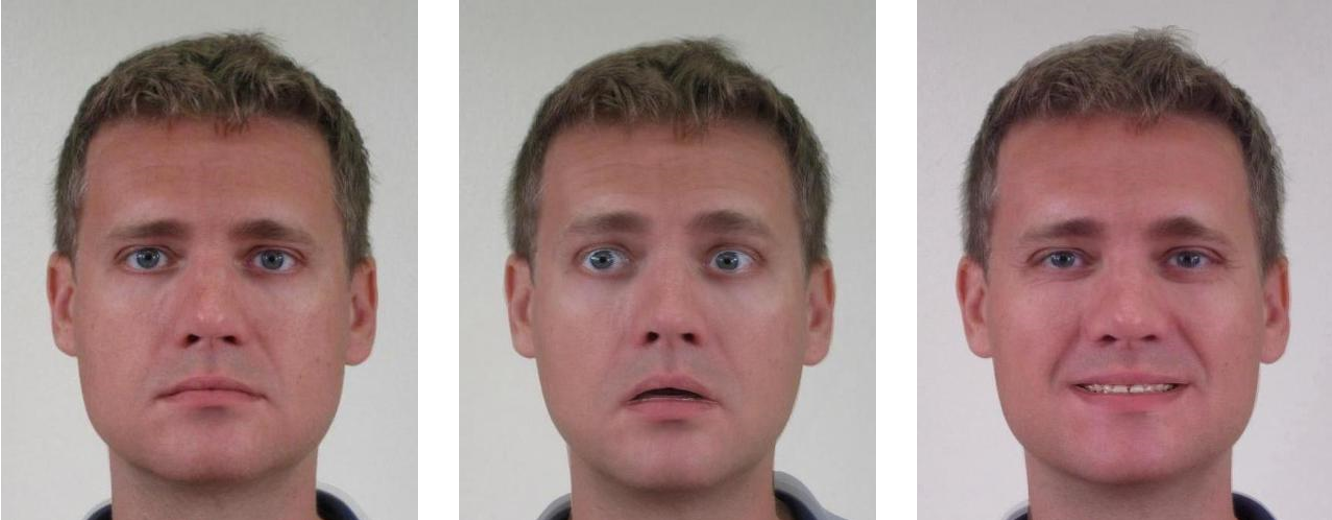}	
	\caption{Different facial avatars,  each avatar representing an algorithmic assistant and not a loan applicant, and  trained to provide useful advice through facial expressions.
	The leftmost avatar is set to a neutral expression ($z=0$).
	\label{fig:avatar}}
\end{figure}

\subsection{Decision-compatible Algorithmic Avatars}

For this experiment we
consider a real decision task and use real data (approving loans),
train with many humans participants (mTurkers),
and explore a novel form of representational advice (facial avatars).
Altogether we elicit around 5,000 human decisions
for training and evaluation.
% \footnote{All experiments are 
% conducted subject to ethical review by the university's IRB.}
Specifically we use the  {\em Lending Club} dataset,
%This dataset contains approximately 890,000 examples of real loans issued on the peer-to-peer Lending Club platform, both completed loans and in progress loans.
focusing on the resolved loans, i.e., loans that were 
paid in full ($y=1$) or defaulted ($y=0$),
and  only using features that would have been available to lenders at loan inception.\footnote{
 https://www.kaggle.com/wendykan/lending-club-loan-data}
The decision task is to determine whether to approve a loan ($\des=1$)
or not ($\des=0$),
and the loss function  we use is $\loss(y,a)=\one{y \neq a}$. 

\paragraph{Goals, Expectations, and Limitations.} \label{sec:turkgoals}
%\todo{use this paragraph to convey the message that
%it is not our intent to build a system that shows turkers faces so that they can predict loans better than a %machine}
%
%
%\todo{add here an explanation of what the faces are / are not and why we use them } \sophie{covered later}
%
%
%
% {\em Loans and mTurkers.}
Whereas professional decision-makers are inclined to exercise their own judgment and  deviate from machine advice \cite{stevenson2019algorithmic, de2020case},
mTurkers are non-experts and are likely to follow machine predictions \cite{lai2019human,yin2019understanding}.\footnote{We only know of Turk experiments where good human performance from algorithmic advice  
	can be attributed to humans accepting the advice of accurate  predictions~\cite{lai2020harnessing}.}
For this reason, the goal of the experiment is {\em not to demonstrate performance superiority over
purely predictive advice}, nor to show that mTurkers can become expert loan officers.
Rather, the goal is 
to show that abstract representations can convey predictive advice in a  way that requires users to deliberate, 
% to make a decision,
and to explore whether humans use learned representations differently than they use 
machine predictions in making decisions.
In Appendix \ref{sec:opt} we further discuss unique challenges encountered when training with mTurkers in the loop.

%\emph{Avatars.}
%\sophie{I don't want to apologize for the avatars under `limitations'?? I moved this into `representations'}
% Our choice of facial avatars follows from our interest in exploring
% abstract forms of representation that appeal directly to human cognition.
% Faces are processed holistically, with intricate and complex dependencies beyond the sum of their parts \cite{richler2009holistic}.
% % and have been experimentally validated as an abstract representation of information
% % (not in the domain of the input objects) \cite{chernoff1973use, lott1990use}.
% Further, faces leverage innate human cognitive capabilities---immediate, effortless, and fairly consistent processing of facial signals \cite{izard1994innate,kanwisher1997fusiform,todorov2008understanding,freeman2016more}, qualities which have led to previous experimental validation as an abstract (not in the domain of the input objects) representation of information
%  \cite{chernoff1973use, lott1990use}.

\begin{figure}[t]
	\centering
	\includegraphics[trim=1cm 0.2cm 2.4cm 1cm,clip,width=1\columnwidth]{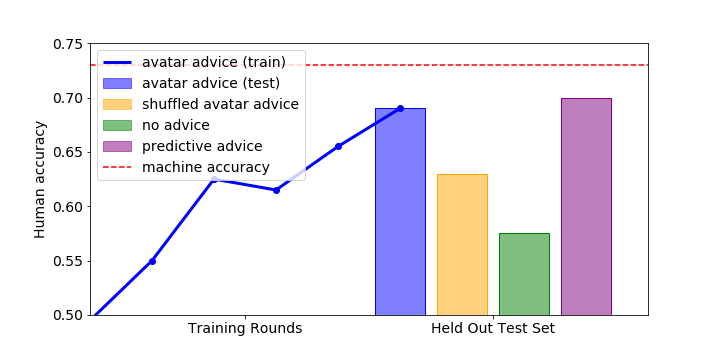}
	\caption{Human accuracy in the algorithmic advice condition (‘avatar advice')
		consistently increases over rounds. Performance quickly surpasses the ‘no advice’ (data only) condition,
		and steadily approaches performance of users observing algorithmic predictions (‘predictive advice’),
		which in itself is lower than machine-only performance (`machine accuracy').
	Human accuracy falls when faces are shuffled within predicted labels of $\hat{h}$,  confirming that faces convey useful, multi-variate information.
	 \label{fig:avatar_acc}}
\end{figure}

\paragraph{Representations.}
% \todo{make reviewers understand that avatars are just like scatterplots}
With the aim of exploring broader forms of representational advice,
 we make use of a \emph{facial avatar}, framed to users as an \emph{algorithmic assistant}--- not the recipient of the loan ---and communicating through its facial expressions
information that is relevant to a loan decision.
The avatar is based on a single, realistic-looking face capable of conveying versatile expressions (Figure~\ref{fig:avatar_acc} includes some examples). % and the Appendix).
Expressions vary along ten dimensions
including {\em basic emotions} \cite{du2014compound},
{\em social dimensions } (e.g., dominance and trustworthiness \cite{du2014compound,todorov2008understanding}),
and subtle changes in {\em appearance } (e.g., eye gaze). % or mild aging effects).
Expressions are encoded by the representation vector $z$,
with each entry corresponding to a different facial dimension.
Thus, vectors $z$ can be thought of as points in $k$-dimensional `face-space' in which expressions vary smoothly with $z$. % that correspond to emotions, social factors, and subtle changes to appearance.
% \sophie{

We are interested in 
facial avatars because they are 
   abstract 
(i.e., not in the domain of the input objects) and
because they have previously been validated as useful  representations of information~\cite{chernoff1973use, lott1990use}. They are also
high-dimensional representations, and non-linear in the input features; that is,
faces are  known to be processed holistically with dependencies beyond the sum of their parts \cite{richler2009holistic}.
% Our choice of facial avatars follows from our 
% Further, this form of representation 
% appeals directly to human cognition.
% Faces are 
% and have been experimentally validated as an abstract representation of information
% (not in the domain of the input objects) \cite{chernoff1973use, lott1990use}.
% Faces 
 Faces also leverage innate human cognition---immediate, effortless, and fairly consistent processing of facial signals \cite{izard1994innate,todorov2008understanding,freeman2016more}.
% kanwisher1997fusiform,
%As high-dimensional representations that are non-linear in the input features, 
%avatars uniquely satisfy the constraints of our problem setting.
% }
%

% \dcp{i think we can drop next sentence, mainly for space and also because it seems unsupported} For a fixed representation mapping, avatars serve to visualize inputs
% in a way that allows humans to infer how a data-point relates
% to other points in the dataset
% (i.e., by making inferences about the facial dimensions).
%
Through \mom, we \emph{learn} a mapping from inputs to avatars
that is useful for decision-making. Training is driven completely by human responses, and
learned expressions reflect usage patterns that users found to be useful, as opposed to hand-coded mappings as in {\em Chernoff faces}~\cite{chernoff1973use}.
% In a successful mapping, the avatars can summarize combinations of variables, but with the range and variation of expression calibrated such that \emph{decision-relevant variation} in the data is salient. In this way, learned avatars can relate data points to the full training set (e.g., encoding how extreme the values are).

\paragraph{Model and Training.}
We set $\repnet$ to be a small, fully connected network with a single 25-hidden unit layer,
mapping inputs to representation vectors $z \in \R^\zdim$.
%(see Appendix \ref{sec:exp3details} for details on model selection).
%mapping $x \mapsto z \in \R^\zdim$ and $z \mapsto \des$, respectively.
The visualization component $\vizmap(z)$ creates avatars 
by morphing a set of base images, each corresponding to a facial dimension,
with $z$ used to weight the importance of each base image.\footnote{
	Morphed images were created using the {\em Webmorph} software package \cite{debruine2016webmorph}.}\footnote{All base images correspond to the same human actor, whose corresponding avatar was used throughout the  experiment.}
For regularization, we additionally consider the loss of a decoder network implemented by an additional neural network, which attempts to reconstruct the input $x$ from the representation.
This term
encourages points in face-space to preserve distances in instance-space at the cost of some reduction in accuracy. 
%(see Appendix \ref{sec:exp3details}).
This promotes representations that carry more information
about inputs than that implied by simple
predictions. 
For $\hhat$ we use a small, fully connected network with two layers of size 20 each,
operating directly on representation vectors $z$.
%dcp cut for space \footnote{\dcpadd{It bears emphasis that we did not need to experiment extensively with different set-ups; and indeed, given that human-in-the-loop training is time-consuming and costly,  this becomes difficult.}} 
%\dcp{discuss somewhere, e.g. footnote here, whethern we experimented mmuch with the phi network and the hhat network}
%

In collecting human decisions for training $\hhat$,
% and in turn learning useful representations,
mTurkers were queried for their decisions regarding the approval or denial of loan applications.\footnote{As all users share the same representation mapping, we restrict to US participants to promote greater cross-user consistency.} % with greater, common cultural understanding of face-space.}
New users were recruited at each round to obtain reports that are as independent as possible and to control for any human learning.
Each user was queried for a random subset of 40 training examples,
with the number of users chosen to ensure that each example would
receive multiple responses (w.h.p.).
For predictive purposes, binary outputs were set to be the majority human response.
Each loan application was presented using the most informative features
as well as the avatar.
% Each example included a subset of features
% informative for machine prediction, along with the avatar.
We did not relate to users any specific way in which they should use avatar advice, and {\em care was taken to ensure users understood that the avatar does not itself represent
% s an algorithmic assistant
% and not 
an applicant}.\footnote{Respondents who did not understand this point in a comprehension quiz were not permitted to complete the task.}
% , consistent with the idea that our framework attempts to model the average human.
Appendix \ref{sec:exp3} provides additional experimental details.

% \sophie{Rather, \mom\ learns from user responses how they would \emph{like} to use the representation, e.g., ``Shifting eye gaze should tell me the participant is more likely to default", and uses these cues to pass along relevant information such as how extreme the defendant's past negative public records are relative to the full  training set (See Results for more discussion)}.
% \nir{I still don't think we have strong enough support to make this sort of claim.}
% \dcp{point to appendix for details on instructions to users in this regard?}

\paragraph{Results.}
Our results show that \mom\ can learn representations that support good decisions 
through a complex, abstract representation, and that this representation 
carries multivariate information, making it  qualitatively different than prediction.
As benchmarks, we consider the accuracy of
a trained neural network model $\nn(x)$ having architecture equal to $\hhat \circ \repnet$
(but otherwise unrelated to our human experiments),
as well as human performance under predictive advice $\advice(x) = \predm \in [0,1]$ 
where $\predm$ is the predicted probability of  $\nn(x)$.
We also consider a condition with `shuffled' avatar advice, which we describe  
below.
% \dcp{next sentence hard to parse. drop?} The data subset described above is present in all conditions.
% In all conditions, participants also see a subset of the original data including all features which are informative for the machine model. 

Figure~\ref{fig:avatar_acc} shows the training process and resulting test accuracy
(data is balanced so chance$\,\,\approx 0.5$).\footnote{Results 
are Statistically significant under one-way ANOVA,
$\text{F}(3,196)=2.98, p<0.03$.}
At first, the (randomly-initialized) representation $\repnet$
produces arbitrary avatars,
and performance in the avatar condition is lower than in the no-advice condition.
This indicates that users take into account the (initially uninformative) algorithmic advice.
As learning progresses, user feedback accumulates
and the accuracy from using  the \mom\ framework steadily rises.
After six rounds, avatar advice contributes to a boost of 11.5\% in accuracy (0.69) over the no-advice condition (0.575),
reaching 99\% of the accuracy in the predictive advice condition (0.70). % (0.642 of 0.68)
Performance in the predictive advice condition does not reach  machine  accuracy (0.73),
showing that not all subjects follow predictive advice.

\paragraph{Analysis.}
We additionally explore what the representations learn, and how humans incorporate them into predictions. One possible concern is that despite regularization,
learned avatars may simply convey stylized binary predictions
% through fancy graphics 
(e.g., happy or sad faces).
To explore this, we added a `shuffled' condition in which faces are shuffled within predicted labels of $\hhat$.
As shown in Figure~\ref{fig:avatar_acc},
shuffling degrades performance, confirming
% and confirms
that faces convey more 
% complex 
information than the system's  binary prediction.
Moreover, the avatars do not encode
% provide a simple encoding of 
a univariate (but not binary) prediction, 
and humans do not use the information in the same way that they use numeric predictions:
(i) no single feature of $z$ has a correlation with predicted human responses $\hhat(z)$ of more than $R^2=0.7$,
% (ii) accuracy of predicting human responses from any single feature is at most 0.695,
% (iii)
(ii) correlations of average human response with features $z$ are low ($R^2 \le 0.36$ across features) while responses in the predictive condition have $R^2=0.73$ with the predictions, and
% (iv)
(iii) users in the avatar condition self-report using the data as much or more than the advice 83\% of the time, compared to 47\% for the predictive advice condition. % (See Appendix \ref{sec:exp3}). 
% We believe this analysis shows promise in the ability for 
% representations such as avatars to convey 
% useful predictive information while mitigating automation bias.

% \todo{Include new analysis here:
% - x predicted from z
% - example of interesting coefficients in x predicted from z
% - example of interesting $r^2$ values of x vs z
% }

At the same time, $z$ preserves important information regarding $x$.
To show this, we train linear models to predict from $z$
each of the data features:
interest rate (\feat{rate}),
loan term (\feat{term}),
debt to income ratio (\feat{dti}),
negative public records (\feat{rec}),
annual income (\feat{inc}),
employment length (\feat{emp}).
Results show that 
$z$ is highly informative of \feat{rate} ($R^2=0.79$) and \feat{term} ($0.57$),
mildly informative of \feat{rec} ($-0.21$),
\feat{inc} ($0.23$), and \feat{emp} ($0.13$),
and has virtually no predictive power of \feat{dti} ($-0.03$).
Further inspecting model coefficients reveals a complex pattern
of how $z$ carries information regarding $x$
(see Appendix \ref{sec:avatarreps} for all coefficients).
E.g.:
% in general sign of coeffs aligns with reasonable interpretation
trustworthiness plays an important part in predicting all features,
whereas anger is virtually unused;
happiness and sadness do not play opposite roles---happiness is significant in \feat{term}, while sadness is significant in \feat{rate};
and whereas \feat{emp} is linked almost exclusively to age variation,
\feat{inc} is expressed by over half of the facial dimensions.

%==================================================================

\subsection{Incorporating Side Information}

To demonstrate additional capabilities of \mom\, we    show that the 
framework can also learn representations that allow a decision maker to 
leverage side information 
that is unavailable to the machine.
%\nir{it reads like zemel also did side info} true, and ok not to say because this is in related work now 
Access to side information is one advantage humans may have over machines,
and our goal here is to show the potential of representations in eliciting decisions whose quality
surpasses that attainable by machines alone.
%
%
% Referencing our earlier discussion of mTurk, 
We adopt simulation  for this experiment  because it is  challenging  for
non-experts (like mTurkers) to outperform purely predictive advice,
even with access to additional side information.
%\footnote{As discussed in the related work, simulated human models have also been used in an experimental study of fairness  by~\citet{zemel2013learning}.}
Simulation also allows us to systematically vary
the synthetic human model, and we consider four distinct models of 
 decision-making.  
 %
% \dcp{if there are other papers that use simulations to study human-machine cooperation, add them here, e.g. in a footnote}

% \if 0
% One  advantage humans can have is access to {\em side information}, 
% relevant to the decision and unavailable to the machine (at training or
% test time). We evaluate \mom\ in this setting,
% and show  it can provide a representation that helps users achieve 
% performance surpassing its own.
% \fi\

\if 0
Our main focus in this experiment is on demonstrating how our learning pipeline can adapt to this setting,
%Accounting for side-information
and as it introduces unique challenges %, both algorithmic on human-related, 
that go beyond those studied in our other experiments,
here we evaluate \mom\ in a synthetic environment and using a human-proxy model
(as in many other current works on human-machine cooperation).
%
\fi
%
% The task 
We consider a medical decision-making task in which doctors must evaluate
the health risk of incoming ER patients and have access to a predictive model. \footnote{MDCalc.com is one example of a  risk assessment calculator for use by medical professionals.} % (www.mdcalc.com).}
%The risk model has access to four binary variables.
%dcp cut space Predictive models are trained on data available in medical records,
%and provide a risk estimate for each patient given her features.
%dcp cut space We assume that doctors have access to the ,
%as well as to the model %(i.e., its coefficients)
%and its predictions.
%
Here, we focus on compact, linear models, and view the model coefficients along with the input features as the representation, affecting the decision process of doctors.
%as well as to the model itself (i.e., the coefficients) and to its predictions,
Doctors additionally have access to side information that is \emph{unavailable to the model} and  may affect their decision.
Our goal is to learn a 
% compact, linear 
model that can account for how doctors
% choose to 
use this side information.

\paragraph{Setup.}
There are four primary binary features $x \in \{0,1\}^4$: diabetes ($x_d$), cardiovascular disease ($x_c$), race ($x_r$), and income level ($x_i$). An integer `side-information' variable $\side \in \{0,1,2,3\}$ encodes how long the patient's condition 
was allowed to 
progress before coming to the ER and is available only to the doctor.
%\nir{can we justify why this is a reasonable choice for side info? is it never collected? is it something doctors assess on the fly?}
We assume ground-truth risk $y$ is determined only by diabetes, cardiovascular disease, and time to ER,
through $y = x_d + x_c + \side$,
where $x_d, x_c, \side$ are sampled independently.
We also assume that $x_r,x_i$ jointly correlate with $y$
(e.g. due to disparities in access),  albeit not perfectly,
 so that they carry some but not all signal in $\side$, whereas $x_d,x_c$ do not; 
see Appendix \ref{sec:side_info_data_gen} for full details).
In this way, $x_r$ and $x_i$ 
% can be useful in prediction
% as they 
offer predictive power beyond that implied by their correlations with known health conditions ($x_d$, $x_c$),
but interfere with use of side information.

We model a decision maker who generally follows predictive advice $\yhat=f_w(x)= \inner{w,x}$,
but with the capacity to adjust the machine-generated risk scores at her discretion
and in a way that depends on the model through its coefficients $w$. We assume that doctors are broadly aware of the correlation structure of the problem,
and are prone to incorporate the available side information $\side$ into $\yhat$
if they believe this will give a better risk estimate.
We model the decisions of a population of doctors as 
incorporating $\side$ additively and with probability that decreases with the magnitude of either of the coefficients $w_r$ or $w_i$.
We refer to this as the \emph{or} model and set
$h_{\text{or}}(x,\side,w)=\yhat+I(w)\cdotp \side$ with $I(w) \propto 1/(\max\{w_r,w_i\})$.
% so that more weight on $w_r$ and $w_i$ reduces the probability of incorporating $s$. %\nir{missing link function}.
We also consider simpler decision models:
\emph{always} using side information ($h_{\text{always}}$),
\emph{never} using side information ($h_{\text{never}}$),
and a \emph{coarse} variant of $h_{\text{or}}$
using binarized side information, 
$h_{\text{coarse}} = \yhat + I(w)\cdotp 2\cdotp\1{\side\geq2}$.

\paragraph{Model.}
The representation $\vizmap(z)$ consists of $x$, coefficients $w$
(these are learned within $\repnet$), and $\yhat = \inner{w,x}$.
\footnote{In an application, the system should convey to users that it is aware they may have side information.}
The difficulty in optimizing $\repnet$ is that $\side$ is never observed,
and our proposed solution is to use $y$ (which is known at train time)
as a proxy for $\side$ when fitting $\hhat$, which is then used to train $\repnet$
(see Section~\ref{sec:method}).
Since $x$ and $y$ jointly carry information regarding $\side$,
we define $\hhat(x,y;w) = \inner{w,x}+\sidehat(x,y)$, where $\sidehat(x,y)=v_0 y + \sum_{j=1}^{4}v_j x_j$,
and $v$ are parameters.
Note
% that $\hhat$ cannot always perfectly reconstruct $h$, and 
that it is enough that $\sidehat$ models how the user {\em  utilizes } side information, rather than the value of  $s$ directly; 
% Because of necessary differences between the differentiable architecture of $\hhat$ and $h$, and 
 $s$ is never observed, and there is no guarantee about the relation between $\sidehat$ and $\side$.

\begin{table}[t]
\centering
\begin{tabular}{lcc}
% \hline
% \textbf{Accuracy} 
& \textbf{\mom} &  $h(\text{\textbf{Machine}})$ \\ \hline
Or                & 1.0                        & .894                \\
Coarse Or         & .951                 & .891               
\\
Never             & .891                   & .891               \\
Always            & 1.0                      & .674               
%  \\ \hline
\end{tabular}
\caption{Performance of \mom\ with side information on four synthetic human models. Machine-only performance is 0.890.
% \mom\ is compared with a machine-only benchmark (no side information) and with the human mapping applied to this machine model.
\label{tab:sideinfo}}
\end{table}

\paragraph{Results.}
We compare \mom\ to two other baselines:
a machine-only linear regression,
and the human model $h$ applied to this machine-only  model,
and evaluate performance on the four  synthetic human models
($h_{\text{or}}, h_{\text{coarse}}, h_{\text{never}}$, and $h_{\text{always}}$).
% (\emph{or}, \emph{coarse}, \emph{always}, and \emph{never})
Both \mom\ and the baselines use a linear model but  the model in \mom\ is trained to
take into account how users incorporate side information.
For evaluation, we consider binarized labels %$y$ such that patients with risk $>3$ have
$y_{bin}=\1{y>3}$.

We report results averaged over ten random data samples of size 1,000
with an 80-20 train-test split.
As Table \ref{tab:sideinfo} shows, 
due to its flexibility in finding a representation that allows for incorporation of side information  by the user,
 \mom\ reaches 100\% accuracy for the  {\em or} and {\em always} decision models.
\mom\ maintains its advantage  under the {\em coarse-or } decision model
(i.e., when doctors use imperfect information),
and remains effective in settings where side information is  never used.
% \dcp{need an explanation for why 100\% accuracy when always used, e.g. footnote}
%
% in which $y$ is a biased estimator of $e$, 
The problem with the baseline model is that it includes non-zero coefficients
for all four features. This promotes accuracy in a machine-only setting, and in the absence of side information. Given this, the \emph{or} and \emph{coarse-or} decision models only very rarely introduce the side information--- and this is indeed the best they can do given that the machine model uses all four variables. In contrast, for the {\em always} decision model the
user always introduces side information, causing over-counting of the time to ER effect on patient outcomes (because of 
correlations between $\side$ and $x_r$ and $x_i$).
In contrast, \mom\ learns a  linear model  that is responsive to the human
decision-maker: for example, including  non-zero coefficients
for only $x_d$ and $x_c$ 
with the {\em or} decision model.

\section{Discussion} \label{sec:discussion}

% \todo{merge with NeurIPS submission version}
% \todo{maybe move things to broader impact section?}

We have introduced a novel learning framework for supporting human decision-making.
%Key to our work is how we consider the role of algorithms:
Rather than  view algorithms as  experts, asked to explain their conclusions to people,
we position algorithms as advisors whose goal is to help humans
make better decisions while retaining human agency.
% agency with people.
The \mom\ framework learns to provide representations of inputs that provide advice and promote good decisions. 
%We demonstrate success in learning to  support human decision models.
%beyond quality control and deferral.
% \dcp{1-2 more sentences to summarize what we have shown}
We see this as a promising direction for promoting synergies between learning systems and people and hope that by tapping into innate cognitive human strengths, % through targeted supervision,
learned representations can improve human-machine collaboration by prioritizing information,
highlighting alternatives, and correcting biases. 

Our hope is that centering humans in the decision process will lead to augmenting intelligence but also facilitate transparency.
Unfortunately, this may not always be the case, and ethical, legal, and societal aspects of systems that are optimized to 
promote particular human decisions
must be subject to scrutiny by both researchers and practitioners.

We believe algorithmic decision support, when thoughtfully deployed, exhibits great potential.
% We see reasons to be optimistic regarding the future of algorithmic decision support. 
Systems designed specifically to provide users with the information and framing they need to make good decisions can harness the strengths of both computer pattern recognition and human judgment and information synthesis. 
We can hope that the combination of mind and machine can do better than either alone.
The ideas presented in this paper serve as a step toward this goal.

We advocate for responsible and transparent  deployment of
models with ``h-hat-like" components, in which system goals and user goals are aligned, and humans are aware of what information they provide about their thought processes. Opportunities and dangers of our framework generally reflect those of the broader field of persuasive technology, and ethical guidelines developed in that community should be carefully considered \cite{fogg1998persuasive, berdichevsky1999toward}.

\bibliography{mom_refs}
\bibliographystyle{icml2021}

\clearpage
\onecolumn
%\makeappendixtitle
\appendix

\begin{appendices}

\section{Optimization Algorithm} \label{sec:algo}

% !TEX root = faceval_nips19.tex

\begin{algorithm}[H]
%	\footnotesize
	\algofontsz
	\caption{{Alternating optimization algorithm}}
	\begin{algorithmic}[1]
		\STATE{Initialize $\theta = \theta_0$}
		%		\STATE{$t \gets 0$}
		\REPEAT
		\STATE{$x_1, \dots, x_n \sim \smplst$} 
		\hfill \COMMENT{Sample $n$ train examples}
		\STATE{$z_i \gets \phi_\theta(x_i) \,\,\, \forall \, i \in [n]$}
		\hfill \COMMENT{Generate representations}
		\STATE{$\des_i \gets h(\vizmap(z_i)) \,\,\, \forall \, i \in [n]$}
		\hfill \COMMENT{Query human decisions}
		\STATE{$\smplstaux =  \{(z_i, \des_i)\}_{i=1}^n$}
		\STATE{$\hhatprm \gets \argmin_{\hhatprm'} \expect{\smplstaux}{\loss(\des,\hhat_{\hhatprm'}(z))}$}
		% 			\STATE{$\eta \gets \argmin_\eta \Loss_{\smplst'}(\hhat_\eta)$}
		\hfill \COMMENT{Train $\hhat$} % on human inputs}
		\STATE{$\theta \gets \argmin_{\theta'} \expect{\smplst}{\loss(y,\hhat_\hhatprm(\repnet_{\repprm'}(x)))}$}
		% 			\STATE{$\theta \gets \argmin_{\theta'} \Loss_\smplst(\hhat_\eta \circ \repnet_{\theta'})$}
		% 			\todo{make sure $\Loss$ is defined before}
		\hfill \COMMENT{Train $\repnet$} % on train set}
		%			\STATE{$t \gets t+1$}
		\UNTIL{convergence}
	\end{algorithmic}
	\label{algo:pingpong}
\end{algorithm}

% \begin{algorithm}[H]
% %	\footnotesize
% 	\algofontsz
% 	\caption{{Alternating optimization algorithm}}
% 	\begin{algorithmic}[1]
% 		\State{Initialize $\theta = \theta_0$}
% 		%		\State{$t \gets 0$}
% 		\Repeat
% 		\State{$x_1, \dots, x_n \sim \smplst$}
% 		\Comment{\cmntfontsz{Sample $n$ train examples}}
% 		\State{$z_i \gets \phi_\theta(x_i) \,\,\, \forall \, i \in [n]$}
% 		\Comment{\cmntfontsz{Generate representations}}
% 		\State{$\des_i \gets h(\vizmap(z_i)) \,\,\, \forall \, i \in [n]$}
% 		\Comment{\cmntfontsz{Query human decisions}}
% 		\State{$\smplstaux =  \{(z_i, \des_i)\}_{i=1}^n$}
% 		\State{$\hhatprm \gets \argmin_{\hhatprm'} \expect{\smplstaux}{\loss(\des,\hhat_{\hhatprm'}(z))}$}
% 		% 			\State{$\eta \gets \argmin_\eta \Loss_{\smplst'}(\hhat_\eta)$}
% 		\Comment{\cmntfontsz{Train $\hhat$}} % on human inputs}
% 		\State{$\theta \gets \argmin_{\theta'} \expect{\smplst}{\loss(y,\hhat_\hhatprm(\repnet_{\repprm'}(x)))}$}
% 		% 			\State{$\theta \gets \argmin_{\theta'} \Loss_\smplst(\hhat_\eta \circ \repnet_{\theta'})$}
% 		% 			\todo{make sure $\Loss$ is defined before}
% 		\Comment{\cmntfontsz{Train $\repnet$}} % on train set}
% 		%			\State{$t \gets t+1$}
% 		\Until{convergence}
% 	\end{algorithmic}
% 	\label{algo:pingpong}
% \end{algorithm}

\section{General Optimization Issues} \label{sec:opt}

\subsection{Initialization}

Because acquiring human labels is expensive, it is important to initialize $\repnet$ to map to a region of the representation space in which there is variation and consistency in human reports, such that gradients lead to progress in subsequent rounds.

In some representation spaces, such as our 2D projections of noisy 3D rotated images, this is likely to be the case (almost any 3D slice will retain some signal from the original 2D image). However, in 4+ dimensions, as well as with the subset selection and avatar tasks, there are no such guarantees.

To minimize non-informative queries, we adopt two initialization strategies:
\begin{enumerate}
    \item \textbf{Initialization with a computer-only model:} In scenarios in which the representation space is a (possibly discrete) subset of input space, such as in subset selection, the initialization problem is to isolate the region of the input space that is important for decision-making. In this situation, it can be useful to initialize with a computer-only classifier. This classifier should share a representation-learning architecture with $\repnet$ but can have any other classifying architecture appended (although simpler is likely better for this purpose). This should result in some $\repnet$ which at least focuses on the features relevant for classification, if not necessarily in a human-interpretable format. 
    \item \textbf{Initialization to a desired distribution with a WGAN:} In scenarios in which the initialization problem is to isolate a region of representation space into which to map all inputs, as in the avatar example, in which we wish to test a variety of expressions without creating expression combinations which will appear overly strange to participants, it can be useful to hand-design a starting distribution over representation space and initialize $\repnet$ with a Wasserstein GAN \cite{arjovsky2017wasserstein}. In this case, we use a Generator Network with the same architecture as $\repnet$ but allow the Discriminator Network to be of any effective architecture. As with the previous example, this results in an $\repnet$ in which the desired distribution is presented to users, but not necessarily in a way that reflects any human intuitive concept.
\end{enumerate}

\subsection{Convergence}

As is true in general of gradient descent algorithms, the \mom\ framework is not guaranteed to find a global optimum but rather is likely to end up at a local optimum dependent on both the initialization of $\repnet$ and $\hat{h}$. In our case, however, the path of gradient descent is also dependent on the inherently stochastic selection and behavior of human users. If users are inconsistent or user groups at different iterations are not drawn from the same behavior distribution, it is possible that learning at one step of the algorithm could result in convergence to a suboptimal distribution for future users. It remains for future work to test how robust machine learning methods might be adapted to this situation to mitigate this issue.

%-------------------------------------------------------
\begin{figure}[th!]
\centering
\subfloat[Initial]{\includegraphics[trim=1cm 2.75cm 1cm 2cm, clip,width=0.31\linewidth]{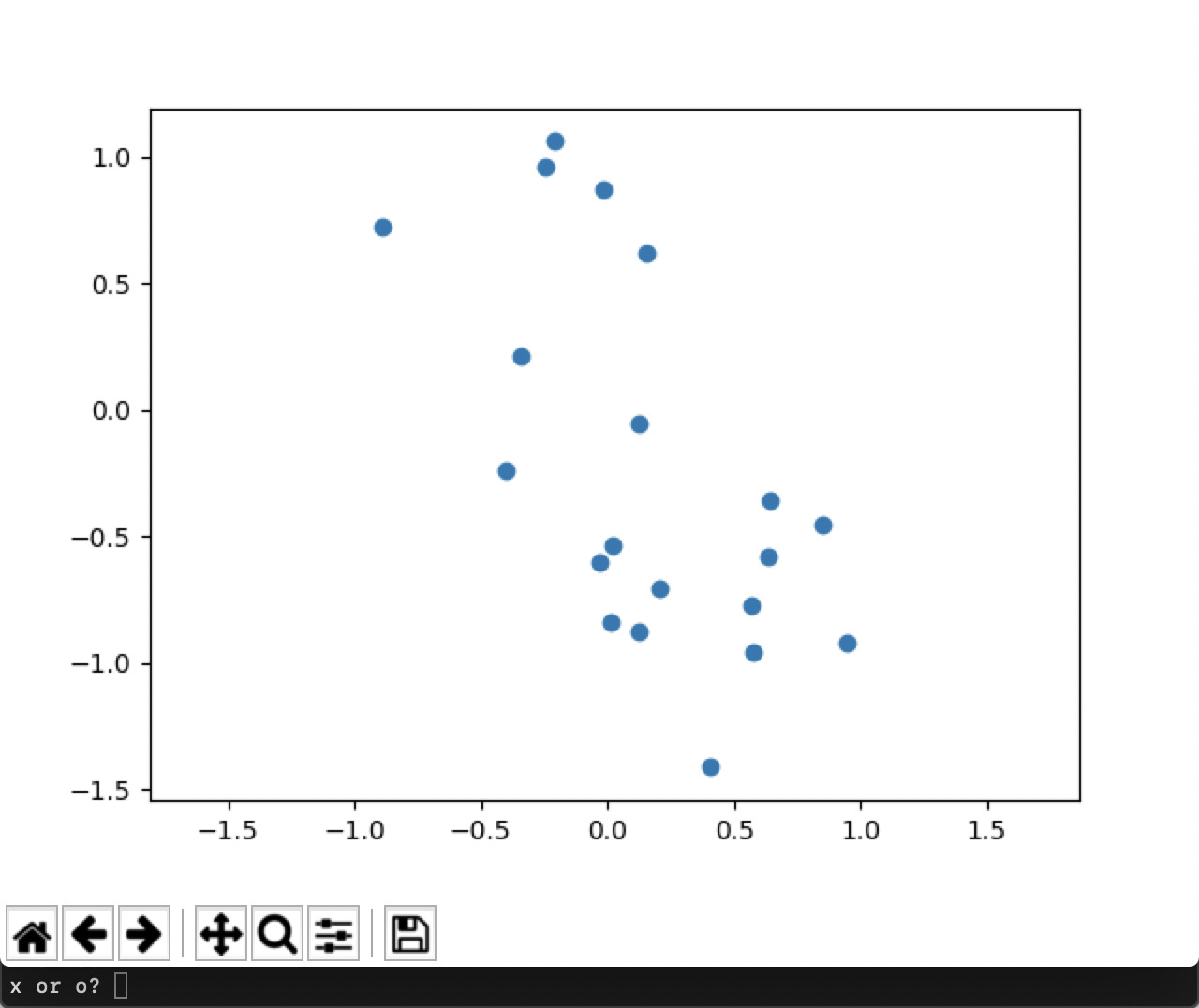}}% 
\quad
\subfloat[Step 3 `x']{\includegraphics[trim=1cm 2.75cm 1cm 2cm, clip,width=0.31\linewidth]{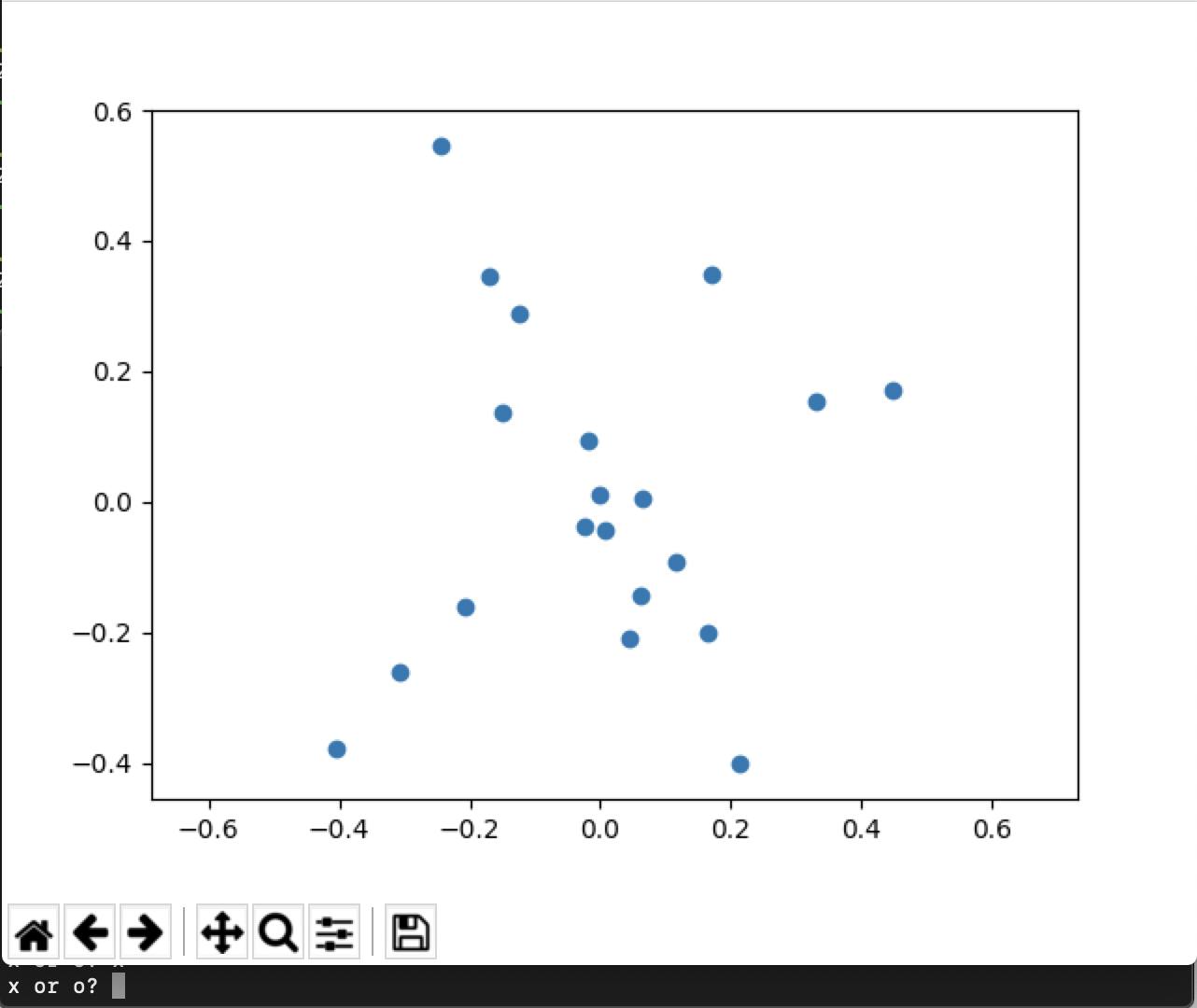}}%
\quad 
\subfloat[Step 4 `x']{\includegraphics[trim=1cm 2.75cm 1cm 2cm, clip,width=0.31\linewidth]{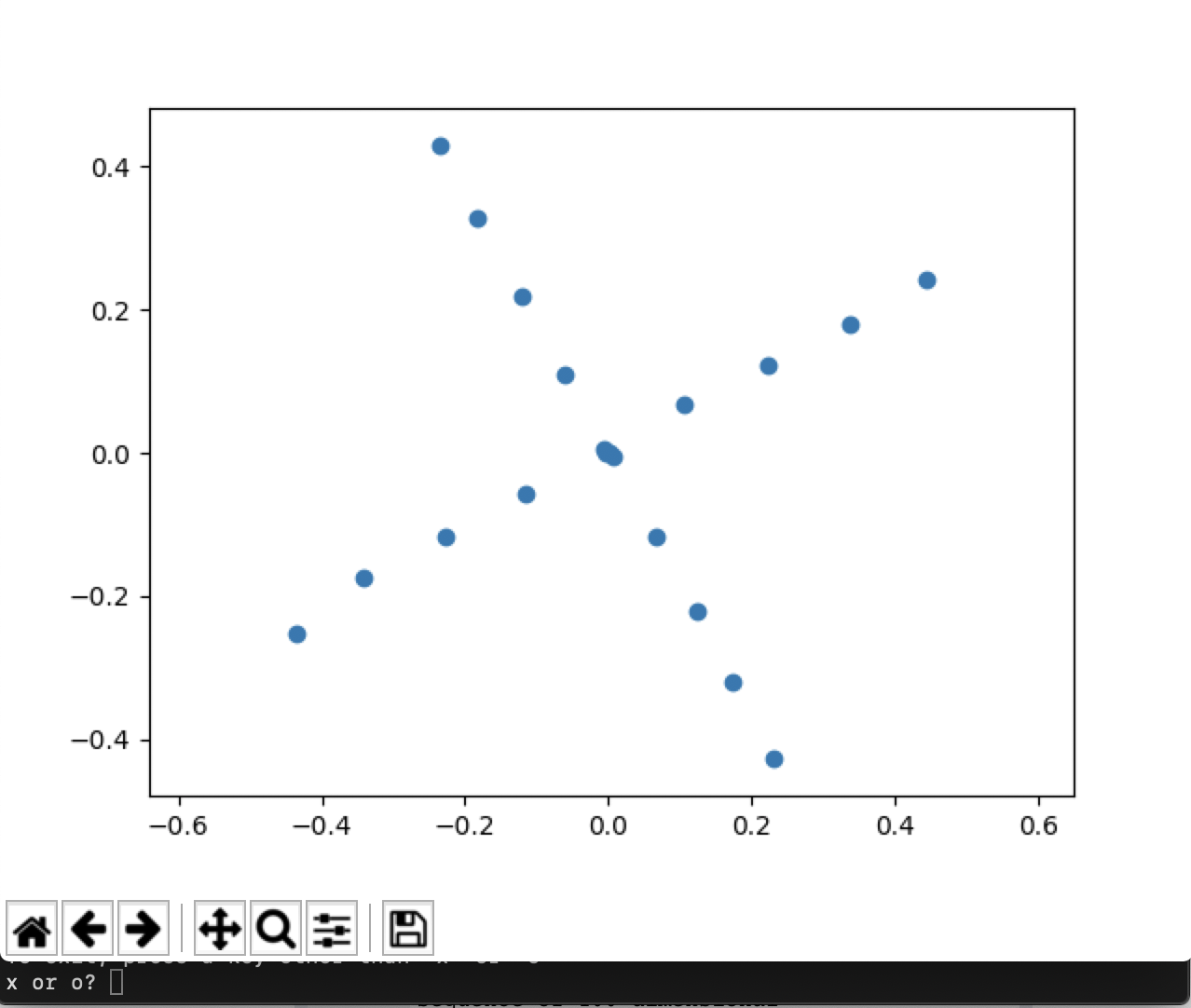}}%
\\
\hspace{5cm}
\subfloat[Step 3 `o']{\includegraphics[trim=1cm 2.75cm 1cm 2cm, clip,width=0.31\linewidth]{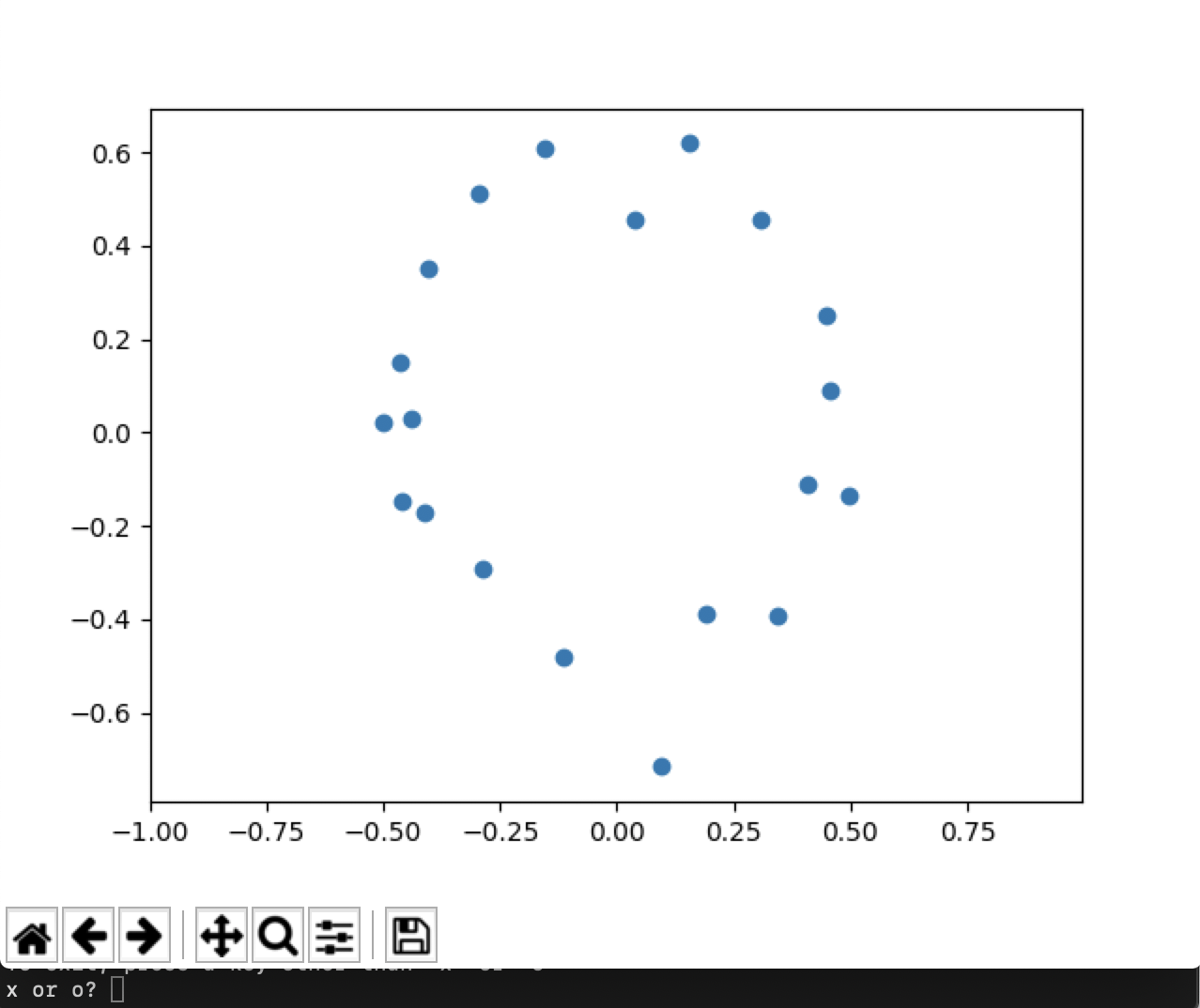}}%
\quad 
\subfloat[Step 4 `o']{\includegraphics[trim=1cm 2.75cm 1cm 2cm, clip,width=0.31\linewidth]{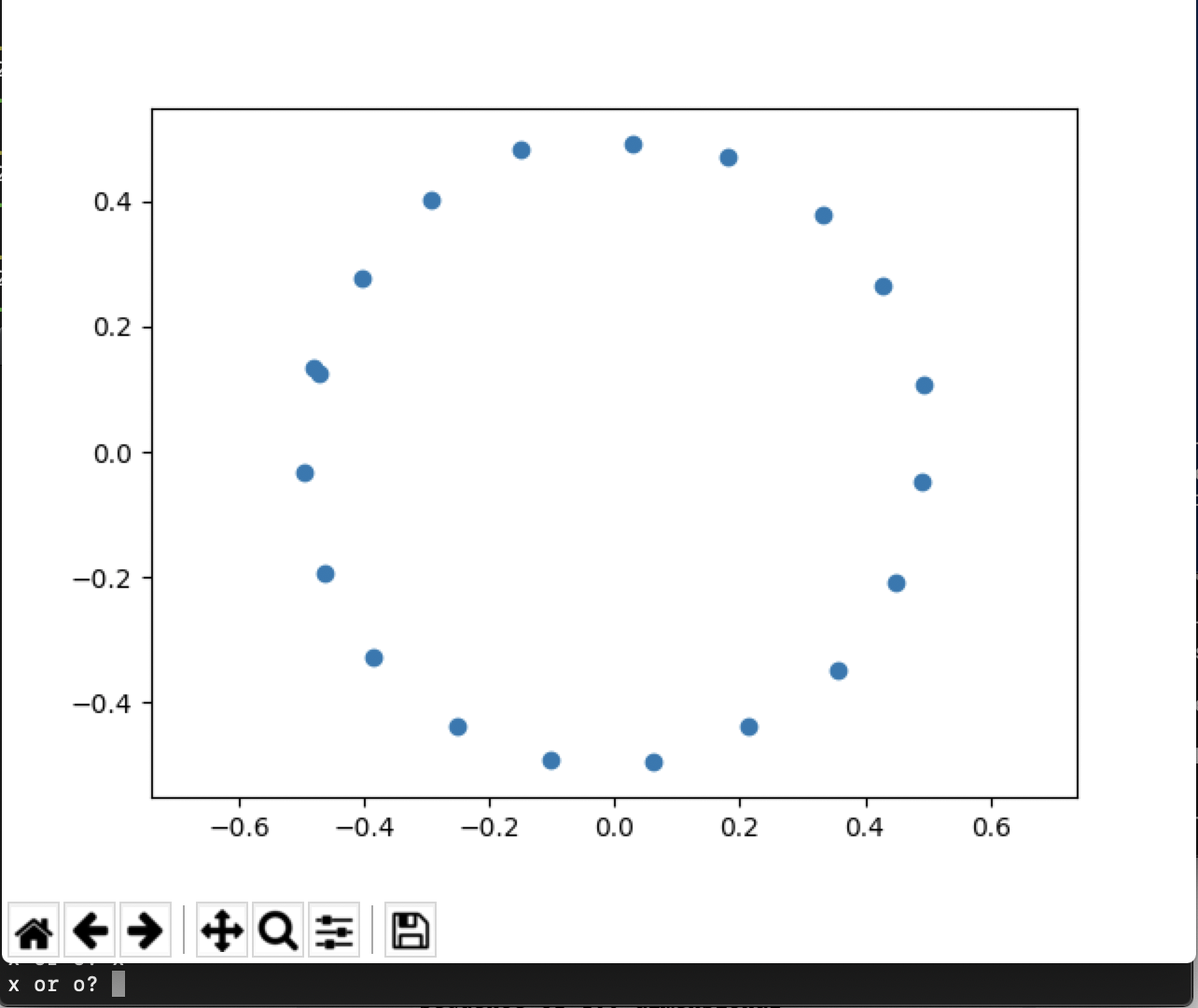}}%
    \caption{Images of x-o interface
	\label{fig:xos} }
\end{figure}
%-------------------------------------------------------

\subsection{Regularization/Early Stopping}

As mentioned in Section \ref{sec:method}, training $\repnet$ will in general shift the distribution of the representation space away from the region on which we have collected labels for $\hat{h}$ in the previous iterations, resulting in increasing uncertainty in the predicted outcomes. We test a variety of methods to account for this, but developing a consistent scheme for choosing how best to maximize the information in human labels remains future work.
\begin{itemize}
    \item \textbf{Regularization of $\hat{h}$:} We test regularization of $\hat{h}$ both with Dropout and L2 regularization, both of which help in preventing overfitting, especially in early stages of training, when the representation distribution is not yet refined. As training progresses and the distribution $\repnet_\repprm(x)$ becomes more tightly defined, decreasing these regularization parameters increases performance.
    \item \textbf{Training $\hat{h}$ with samples from previous iterations}: We also found it helpful in early training iterations to reuse samples from the previous human labeling round in training $\hat{h}$, as inspired by [Bobu et al. 2018]. \footnote{Bobu, Andreea, et al. "Adapting to continuously shifting domains." (2018).}
    % \cite{bobu2018adapting}.
    We weight these samples equally and use only the previous round, but it may be reasonable in other applications to alter the weighting scheme and number of rounds used.
    \item \textbf{Early stopping based on Bayesian Linear Regression:} In an attempt to quantify how the prediction uncertainly changes as $\repprm$ changes, we also implement Bayesian Linear Regression, found in [Riquelme et al., 2018] \footnote{Riquelme, Carlos, George Tucker, and Jasper Snoek. "Deep bayesian bandits showdown." \emph{International Conference on Learning Representations}. 2018.}
    % \cite{riquelme2018deep}
    to be a simple but effective measure of uncertainty, over the last layer of $\hat{h}(\repnet_\repprm)$ as we vary $\repprm$ through training. We find that in early iterations of training, this can be an effective stopping criterion for training of $\repnet$. Again, as training progresses, we find that this mostly indicates only small changes in model uncertainty. 
\end{itemize}

\subsection{Human Input}

Testing on mTurk presents various challenges for testing the \mom\ framework:
\begin{itemize}
    \item 
    In some applications, such as loan approval, mTurk users are not experts.
   This makes it difficult to convince them that anything is at stake (we found that bonuses did not meaningfully affect performance).
    It is also difficult to directly measure effort, agency, trust, or autonomy,
    all of which result in higher variance in responses.
    \item    In many other applications, the ground truth 
    is generated by humans to begin with (for example, sentiment analysis).
    Since we require ground truth for training,
    in these task it cannot be expected of humans to outperform machines.
    \item
    As the researchers found in \cite{lage2018human}, there can be a large variance in the time users take to complete a given task. Researchers have found that around 25\% of mTurk users complete several tasks at once or take breaks during HITs [Moss and Litman, 2019].\footnote{A. J. Moss and L. Litman. How do most mturk workers work?, Mar 2019.}
    % \cite{moss_litman_2019},
    making it difficult to determine how closely Turkers are paying attention to a given task. We use requirements of HIT approval rate greater than 98\%, US only, and at least 5,000 HITs approved, as well as a simple comprehension check. 
    \item
    Turker populations can vary over time and within time periods,
    again leading to highly variable responses, which can considerably
    effect the performance of learning.
    \item    Recently, there have been concerns regarding the usage of automated bots
    within the mTurk communiy.
    Towards this end, we incorporated in the experimental survey
    a required reading comprehension task and as well as a CAPTCHA task,
    and filtered users that did not succeed in these.
\end{itemize}

% in that the users are not experts\nir{this is true in some tasks (like loans), but in others (like text sentiment) its exactly the opposite problem - labels are generated by humans, and the machine's goal is to match it, so there's no chance of humans improving on top of that}, it is difficult to convince them that anything is at stake (we found that bonuses did not meaningfully affect performance), and it is difficult to measure effort, all of which result in higher variance in responses. \nir{these should be bullet points. also turk population might change over time.}

% \nir{and captcha!}

% \nir{maybe add a subsection on smart/active selection of points to query $h$ on?}

% \todo{add footnote on what happens when there are multiple users}

\section{Experimental Details}

% \dcp{figure with x-o images needs to be bigger}

\subsection{Decision-compatible 2D projections} \label{sec:exp1}

In the experiment, we generate 1,000 examples of these point clouds in 3D.
The class of $\repnet$ is a 3x3 linear layer with no bias, where we add a penalization term on $\repnet^{T}\repnet - \mathbb{I}$ during training to constrain the matrix to be orthogonal. Humans are shown the result of passing the points through this layer and projecting onto the first two dimensions. The class of $\hat{h}$ is a small network with 1 3x3 convolutional layer creating 3 channels, 2x2 max pooling, and a sigmoid over a final linear layer. The input to this network is a soft (differentiable) 6x6 histogram over the 2D projection shown to the human user.

We tested an interactive command line query and response game on 12 computer science students recruited on Slack and email. Users filled out a consent form online, watched an instructional video, and then completed a training and testing round, each with up to 5 rounds of 15 responses. Due to the nature of the training process, achieving 100\% accuracy results in $\repnet$ not updating in the following round. With this in mind, if a user reached 100\% accuracy in training, they immediately progressed to testing. If a user reached 100\% accuracy in testing, the program exited.  $\repnet$ was able to find a representation that allowed for 100\% accuracy 75\% of the time, with an average 5 round improvement of 23\% across all participants. Many times the resulting projection appeared to be an `x' and `o', as in Figure \ref{fig:xos}, but occasionally it was user-specific. For example, a user who associates straight lines with the `x' may train the network to learn any projection for `x' that includes many points along a straight line.

The architecture of $\repnet$ and $\hat{h}$ are described in Section \ref{sec:experiments}. For training, we use a fixed number of epochs (500 for $\hat{h}$ and 300 for $\repnet$) with base learning rates of .07 and .03, respectively, that increase with lower accuracy scores and decrease with each iteration. 
We have found these parameters to work well in practice,
but observed that results were not sensitive to their selection.
% These parameters are not carefully chosen, as they were sufficient to make the task work in practice.
The interface allows the number of rounds and examples to be determined by the user, but often 100\% accuracy can be achieved after about 5 rounds of 15 examples each.

\subsection{Decision-compatible algorithmic avatars} \label{sec:exp3}

\subsubsection{Data Preprocessing.}
We use the {\em Lending Club} dataset,
which we filter to include only loans for which we know the resolution (either default or paid in full, not loans currently in progress) and to remove all features that would not have been available at funding time. We additionally drop loans that were paid off in a single lump sum payment of at least 5 times the normal installment. This results in a dataset that is 49\% defaulted and 51\% repaid loans. Categorical features are transformed to one-hot  variables. There are roughly 95,000 examples remaining in this dataset, of which we split 20\% into the test set. 

\begin{figure}[t!]
	\centering
	\includegraphics[width=0.35\linewidth]{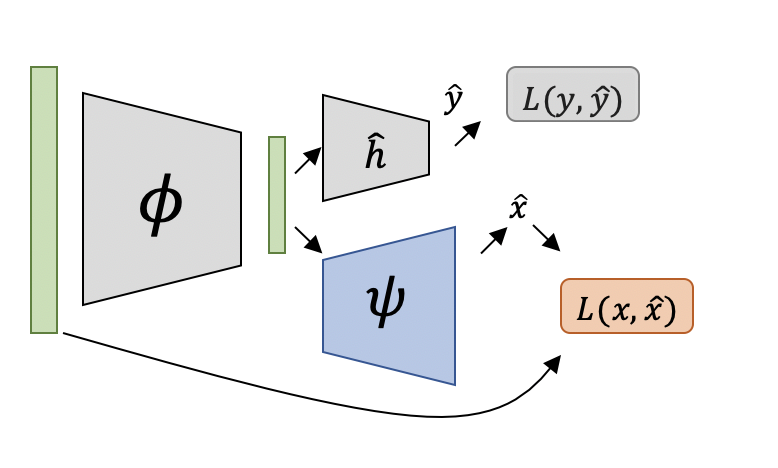}
	\caption{Visualization of reconstruction component}
	\label{fig:recon} 
\end{figure}

\subsubsection{Learning architecture and pipeline.} \label{sec:exp3details}

% \dcp{can we justify why 25? e.g., did we do cross validation?}
% \sophie{this was chosen via CV to be the smallest size capable of recreating a wide distribution of zs with low loss in the WGAN, but recommend to say this in the appendix rather than here}\dcp{sure, can point to app for details}

%  \dcp{can we justify why 20x20, eg.,
% did we try other sizes?}
% \sophie{again, I believe this was chosen from a few different options to be the minimal architecture capable of getting low error in the computer-only version} \dcp{point to details in app}

The network $\repnet$ takes as input the standardized loan data.
Although the number of output dimension are $ \R^\zdim$,
$\repnet$ outputs vectors in  $\R^{11}$. This is because the some facial expressions do not naturally coexist as compound emotions, i.e., happiness and sadness [Du et al., 2014]. \footnote{Shichuan Du, Yong Tao, and Aleix M Martinez.  Compound facial expressions of emotion. \emph{ Proceedings of the National Academy of Sciences}, 111(15):E1454–E1462, 2014.}
% \cite{du2014compound}.
Hence, we must add some additional constraints to the output space, encoded in the extra dimensions. For example, happiness and sadness are split into two separate parameters (rather than using one dimension with positive for happiness and negative for sadness). The same is true of ``happy surprise", which is only allowed to coincide with happiness, as opposed to ``sad surprise." For parameters which have positive and negative versions, we use a tanh function as the final nonlinearity, and for parameters which are positive only, we use a sigmoid function as the final nonlinearity. 

These parameters are programmatically mapped to a series of Webmorph \cite{debruine2016webmorph} transformation text files, which are manually loaded into the batch transform/batch edit functions of Webmorph. We use base emotion images from the CFEE database [Du et al., 2014]
% \cite{du2014compound}
and trait identities from [Oosterhof and Todorov, 2008].\footnote{Nikolaas N Oosterhof and Alexander Todorov.  The functional basis of face evaluation. \emph{Proceedings of the National Academy of Sciences}, 105(32):11087–11092, 2008.}
% \cite{oosterhof2008functional}.
This forms $\vizmap$ for this experiment. 

% \red{we primarily use base emotion images from the CFEE database \cite{du2014compound} and morph them  using Webmorph \cite{debruine2016webmorph}}

% \nir{missing description of hhat}

The network $\repnet$ is initialized with a WGAN to match a distribution of parameters chosen to 
output a fairly uniform distribution of feasible faces.
To achieve this, each parameter was chosen to be distributed according to one of the following: a clipped $\mathcal{N}(0,4)$, $\mathcal{U}[0,1]$, or Beta(1,2).
The choice of distribution was based on inspection as to what would give reasonable coverage over the set of emotional representations we were interested in testing. In this initial version of $\repnet$, $x$ values end up mapped randomly to representations, as the WGAN has no objective other than distribution matching.

\begin{figure}[t!]
\centering
\subfloat[Loss in training $\hat{h}$ over 3 rounds ]{\includegraphics[width=0.4\linewidth]{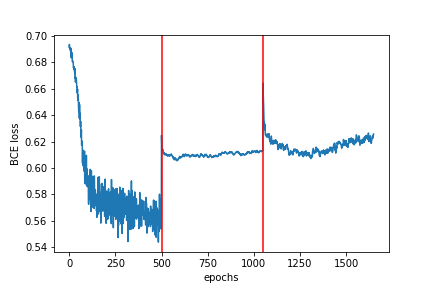}}% 
\qquad \subfloat[Validation Accuracy in training $\repnet$ over 3 rounds]{\includegraphics[width=0.4\linewidth]{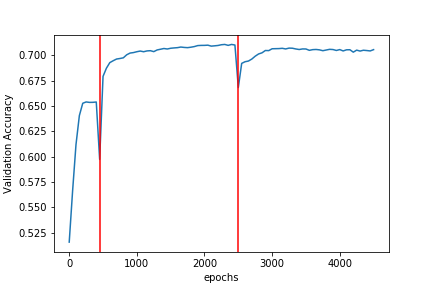}}%
\caption{$\hat{h}$ does not necessarily have to match $h$ well to lead to an increase in accuracy}
	\label{fig:hhat} 
\end{figure}

The hidden layer sizes of $\repnet$ and $\hhat$ were chosen via cross validation. For $\repnet$, we use the smallest architecture out of those tested capable of recreating a wide distribution of representations $z$ as the generator of the WGAN. For $\hhat$, we use the smallest architecture out of those tested that achieves low error both in the computer-only simulation and with the first round of human responses.

\begin{figure}[b!]
\centering
\subfloat[]{\includegraphics[width=0.42\linewidth]{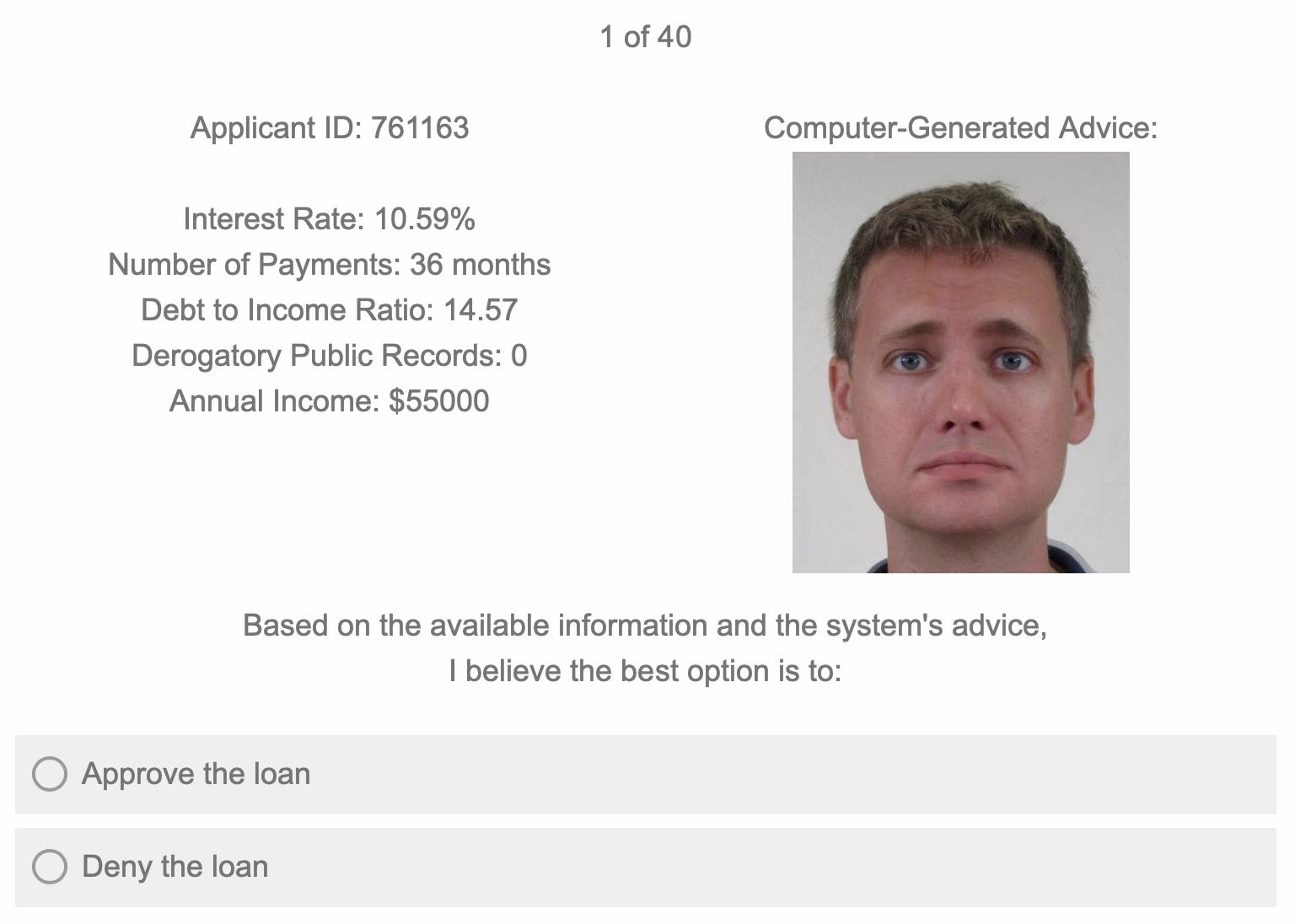}}% 
\quad
\subfloat[]{\includegraphics[width=0.42\linewidth]{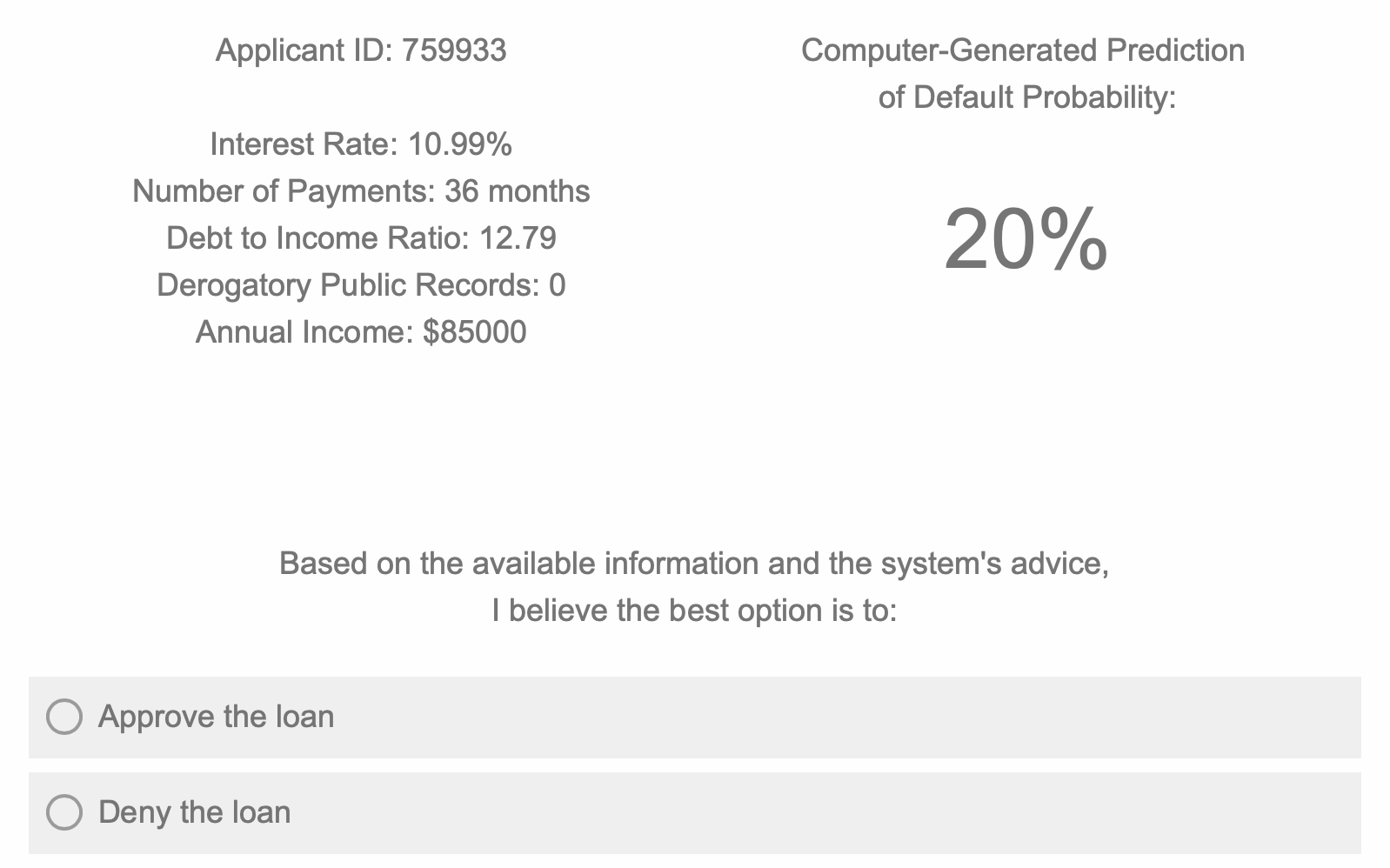}}%
\caption{Images from mTurk questionnaire}
	\label{fig:qualtr_ques} 
\end{figure}

In the first experiment, we collect approximately 5 labels each (with minor variation due to a few mTurk users dropping out mid-experiment) for the LASSO feature subset of 400 training set $x$ points and their $\repnet_0$ mappings (see Figure \ref{fig:qualtr_ques}). $a$ is taken to be the percentage of users responding ``approve" for each point.

\begin{figure}[t!]
\centering
\subfloat[Training Rounds (`Overall` here is average \emph{per user} score, rather than the score of the average response per question)]{\includegraphics[width=0.4\linewidth]{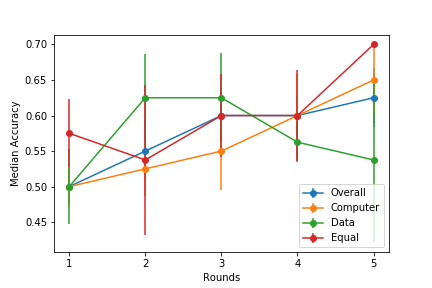}}%
\qquad \subfloat[Test Round]{\includegraphics[width=0.4\linewidth]{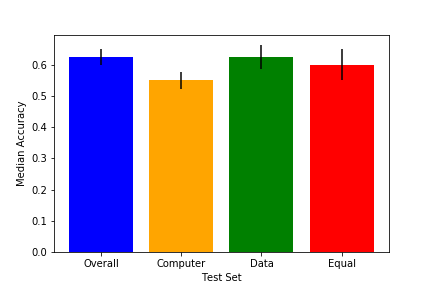}}%
\caption{Results by Reported User Type}
	\label{fig:qualtr} 
\end{figure}

% \david{give the specific on use of LASSO and 5 top features in appendix} 

To train $\hat{h}$, we generate 15 different training-test splits of the collected $\{z,a\}$ pairs and compare the performance of variations of $\hat{h}$ in which it is either initialized randomly or with the $\hat{h}$ from the previous iteration, trained with or without adding the samples from the previous iteration, and ranging over different regularization parameters. We choose the training parameters and number of training epochs which result in the lowest average error across the 15 random splits. In the case of random initialization, we choose the best out of 30 random seeds over the 15 splits.

% \textcolor{red}{To prevent mode collapse, wherein faces ``binarize" to two prototypical exemplars,
% we add a reconstruction regularization term
% $\reg(x) = \|x-\decoder(\repnet(x))\|_2^2$
% to the objective, where $\decoder$ is a decoder implemented by an additional neural network.}

To train $\repnet$, we fix $\hat{h}$ and use batches of 30,000 samples per epoch from the training set, which has 75,933 examples in total. 
% In addition to the reconstruction regularization term
% $\|x-\decoder(\repnet(x))\|_2^2$ 
To prevent mode collapse, wherein faces ``binarize" to two prototypical exemplars,
we add a reconstruction regularization term
$\reg(x) = \|x-\decoder(\repnet(x))\|_2^2$
to the binary cross entropy accuracy loss, where $\decoder$ is a decoder implemented by an additional neural network
(see Figure \ref{fig:recon}).
$\repnet$ here also features a constraint penalty that prevents co-occurrence of incompatible emotions. 

We train $\repnet$ for 2,000 epochs with the Adam optimizer for a variety of values of $\alpha$, where we use $\alpha$ to balance reconstruction and accuracy loss in the form $\mathcal{L}_{total} = \alpha\mathcal{L}_{acc} + (1-\alpha)\mathcal{L}_{rec}$. We choose the value of $\alpha$ per round that optimally retains $x$ information while promoting accuracy by inspecting the accuracy vs. reconstruction MSE curve.
% \nir{this should be made as exact as possible. also explicitly say $\alpha$ is chosen per round.}
We then perform Bayesian Linear Regression over the final layer of the current $\hat{h}$ for every 50th epoch of $\repnet$ training and select the number of epochs to use by the minimum of either 2,000 epochs or the epoch at which accuracy uncertainty has doubled. In all but the first step, this resulted in using 2,000 epochs. 
%
% \nir{did we really use uncertainty? if yes, then we need to remind readers how we measured uncertainty here}
At each of the 2-5th epochs, we choose only 200 training points to query. In the 6th epoch we use 200 points from the test set.

% \todo{add to appendix: regularization, selection of $\lambda$, elbow; see commented-out text below}

\subsubsection{Self-reported user type.}
In the end of the survey, we ask users to report their decision method from among the following choices:
\begin{itemize}
    \item I primarily relied on the data available
    \item I used the available data unless I had a strong feeling about the advice of the computer system
    \item I used both the available data and the advice of the computer system equally
    \item I used the advice of the computer system unless I had a strong feeling about the available data
    \item I primarily relied on the advice of the computer system
    \item Other
\end{itemize}

The percentage of users in each of these groups varied widely from round to round. 

We consider the first two conditions to be the `Data' group, the third to be the `Equal' group, and the next two to be the `Computer Advice' group. 
% While the groups are too small to draw many conclusions from this data, we find that users who report only or primarily using the data increase in mean accuracy from .51 in round 1 to .65 in round 6 ($p<.001$). Note that these accuracies are \emph{per user} rather than the accuracy of the average response as in \ref{sec:experiments} 
Although the trend is not statistically significant (at $p=0.05$), likely due to the small number of subjects per type per round, 
 we find it interesting that the performance improved on average over training rounds for all
three types, of which the equal-consideration type performed best.
For the data-inclined users,
whose performance improved to surpass that of the no-advice condition in as early as
round two,
% , this indicates that the mere presence of facial advice has an affect
% on decisions, in line with previous findings in psychology on the effects of facial exposure \tocitec{hassin}.
this implies at least one of the following: users misreport their decision method; users believe they are not influenced by the advice but   are in fact influenced; 
or, as the algorithmic evidence becomes apparently better, only the population of users who are comparatively skilled at using the data continue to do so.

\subsubsection{Diversity in avatar representation.} \label{sec:avatarreps}

Figure
%====================================
10 \stepcounter{figure}
%====================================
% \ref{fig:avatar_diversity}
%====================================
presents examples of visualized avatars.
Avatars correspond to examples having either low or high 
human-predicted probability (averaged across users) (top figure),
and either low or high machine-predicted probability (lower figure).
For visualization purposes, avatars are aligned according to a uni-dimensional PCA projection of the inputs,
so that their spatial positioning captures the variance in the data.
As can be seen, avatars are different for each predictive category
(positive or negative; human or machine),
but also vary considerably within each predictive category,
with variance eminent across multiple facial dimensions.

We believe the additional dimensionality of the avatar representation relative to a numerical or binary prediction of default is useful for two reasons. Most importantly, high dimensionality allows users to retain an ability to reason about their decisions. In particular, avatars are useful because people likely have shared, mental reference points for faces. 
Moreover, users with a more sophisticated mental reference space  may be able to teach the advising system over time to match specific reasoning patterns to specific characteristics. Additionally, when the advising system does not have a strong conviction about a prediction, presenting neutral advice should encourage the user to revisit the data, whereas percentages above or below the  base rate of default (or 50\%) may suffer from the anchoring effect.

\subsubsection{Further Details on Information Learned by $z$.}
Using cross-validated ridge regression to predict individual $x$ variables from individual $z$ variables results in the coefficients of determination $R^2$ (to 2 significant figures) shown in Table \ref{tab:xonzr2}. %, with abs(coefficients) $>.2$ in bold.

Using cross-validated ridge regression to predict individual $x$ variables from all $z$ variables (both standardized to mean 0, std 1) results in the  \emph{variable coefficients} (to 2 significant figures) shown in Table~\ref{tab:xonzcoef}.

\includepdf{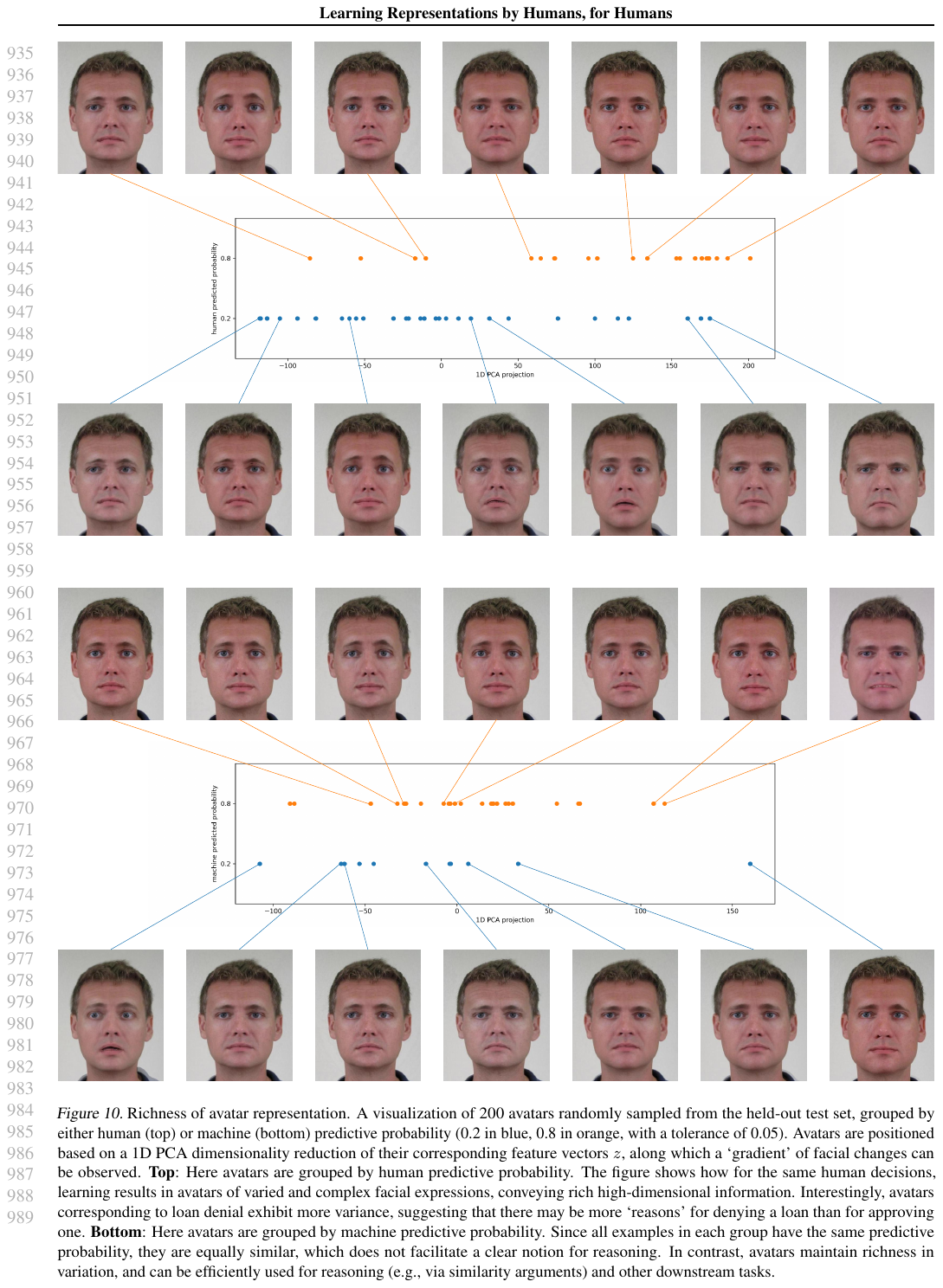}

\subsection{Incorporating Side Information} \label{sec:exp2}

\subsubsection{Data Generation.} \label{sec:side_info_data_gen}
\begin{figure}[t!]
\centering
\begin{tikzpicture}[
roundnode/.style={circle, draw=black!60, very thick, minimum size=7mm},
]
%Nodes
\node[roundnode]      (x_i)                              {$x_i$};
\node[roundnode]        (x_r)       [below=20pt, at=(x_i.south)] {$x_r$};
\node[roundnode]      (x_c)       [right=30pt, at=(x_i.east)] {$x_c$};
\node[roundnode]        (x_d)       [right=30pt, at=(x_r.east)] {$x_d$};
\node[roundnode]        (ER)       [below=20pt, at=(x_d.south)] {$ER$};
\node[roundnode]        (y)       [right=30pt, at=(x_d.east)] {$y$};
% \node[roundnode]      (x_i)                              {$x_i$};
% \node[roundnode]        (x_r)       [below=of x_i] {$x_r$};
% \node[roundnode]      (x_c)       [right=of x_i] {$x_c$};
% \node[roundnode]        (x_d)       [right=of x_r] {$x_d$};
% \node[roundnode]        (ER)       [below=of x_d] {$ER$};
% \node[roundnode]        (y)       [right=of x_d] {$y$};

%Lines
\draw[->] (x_i.east) -- (x_c.west);
\draw[->] (x_r.east) -- (x_d.west);
\draw[->] (x_i.south) -- (ER.north);
\draw[->] (x_r.south) 
.. controls +(down:7mm) and +(left:7mm) .. (ER.west);
\draw[->] (x_c.east) -- (y.north);
\draw[->] (x_d.east) -- (y.west);
\draw[->] (ER.east) -- (y.south);
\end{tikzpicture}
\caption{Relationship of variable correlations in the side information experiment}
\label{fig:dag}
\end{figure}
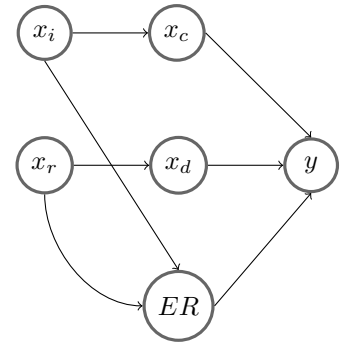

A directed graph showing the variable correlations is shown in Figure \ref{fig:dag}.
The data in the side-information experiment is generated as follows:
A latent variable $l_0 \sim \mathcal{N}(.3, .1)$ introduces a low correlation between $x_i$ and $x_r$ by setting a common mean for their Bernoulli probabilities $l_1, l_2$:
\begin{itemize}
\item $l_1, l_2 \sim \textrm{Unif}(\textrm{max}(l_0 - .3, 0), \textrm{min}(l_0 + .3, 1))$
\item $x_i \sim \textrm{Bernoulli}(1-l_1)$
\item $x_r \sim \textrm{Bernoulli}(1-l_2)$
\end{itemize}

An additional latent variable $l_3$ provides a similar correlation between $x_c$ and $x_d$, which also correlate, respectively, with $x_i$ and $x_r$:
\begin{itemize}
\item $l_3 \sim \textrm{Unif}(.5, .7)$
\item $x_c \sim \textrm{Bernoulli}(l_3 + x_i)$
\item $x_d \sim \textrm{Bernoulli}(l_3 + x_r)$
\end{itemize}

Side information $s$ is highly correlated with $x_r$ and $x_i$ but noisy: $s$ is drawn from a normal distribution centered at $x_r + x_i$ before rounding to an integer value between 0 and 3.
\begin{itemize}
    \item $s_{cont} \sim \mathcal{N}(x_r + x_i, .5)$
    \item $s = \textrm{max}(0,\textrm{min}(3, \textrm{round}(s_{cont})))$
\end{itemize}

% \dcp{explain following a bit more. i think it modifies the previous eqn, and is not about the next eqn?}
The integer outcome variable $y$ is the sum of $x_c$, $x_d$, and $s$. The binary outcome variable $y_{bin}$ is thresholded at $y>3$.

% \dcp{explain following}
$y = x_c + x_d + s$; $;\ y_{bin} = \1{y>3}$

% This data generation process reflects only one possible choice of parameters that attained an appropriate correlation structure.  \dcp{Consider something like `We did experiment with other kinds of set-ups.`}

% We generate datasets with
% $x \in \{0,1\}^{12}$,
% $s \in [0,1,2]$ as follows:
% $\x_1 \hdots x_6 \sim \textrm{Bernoulli}(0.5)$,
% $s \sim \textrm{Cat}([0,1,2], [\frac{1}{3},\frac{1}{3},\frac{1}{3}])$,
% $y = \sum x_1 \hdots x_6 + s$.
% To create a noisy version of $y$ in $x_7 \hdots x_{12}$, we first draw $\x_7 \hdots x_{12} \sim \textrm{Bernoulli}(0.5)$ and then randomly flip entries until $\sum_{i=7}^{12} \x_i = y$. We then add noise by randomly flipping entries of $x_7 \hdots x_{12}$ with probability 0.05. The order of the rows is then randomly permuted such that the relevant rows are in different positions.

%------------------------------------------------------------------------
\begin{table}[b!]
\centering
\caption{Coefficients of Determination $R^2$, predicting each $x$ variable from each final $z$ variable.
% \mom\ is compared with a machine-only benchmark (no side information) and with the human mapping applied to this machine model.
\label{tab:xonzr2}}
\tblfontsz
\begin{tabular}{lcccccc}
  & \feat{rate} & \feat{term} & \feat{dt} & \feat{rec} & \feat{inc} & \feat{emp} \bigstrut[b]\\
\cline{2-7}happiness & 0.00 & \cellcolor[rgb]{ 1,  .949,  .8}-0.15 & \cellcolor[rgb]{ 1,  .949,  .8}-0.14 & 0.00 & -0.01 & 0.00 \bigstrut[t]\\
sadness & -0.01 & -0.06 & -0.10 & 0.00 & -0.04 & -0.07 \\
trustworthiness & \cellcolor[rgb]{ .514,  .235,  .047}\textcolor[rgb]{ 1,  1,  1}{0.57} & \cellcolor[rgb]{ 1,  .949,  .8}0.17 & 0.01 & 0.00 & -0.01 & -0.01 \\
dominance & 0.00 & -0.01 & 0.03 & -0.01 & 0.01 & -0.01 \\
hue & \cellcolor[rgb]{ .514,  .235,  .047}\textcolor[rgb]{ 1,  1,  1}{0.48} & \cellcolor[rgb]{ .957,  .69,  .518}0.29 & -0.02 & 0.00 & -0.04 & -0.02 \\
eye gaze & \cellcolor[rgb]{ .514,  .235,  .047}\textcolor[rgb]{ 1,  1,  1}{0.42} & \cellcolor[rgb]{ .514,  .235,  .047}\textcolor[rgb]{ 1,  1,  1}{0.46} & -0.04 & \cellcolor[rgb]{ .514,  .235,  .047}\textcolor[rgb]{ 1,  1,  1}{-0.40} & -0.04 & \cellcolor[rgb]{ 1,  .949,  .8}-0.17 \\
age & \cellcolor[rgb]{ .957,  .69,  .518}0.23 & \cellcolor[rgb]{ .957,  .69,  .518}0.22 & \cellcolor[rgb]{ 1,  .949,  .8}-0.12 & \cellcolor[rgb]{ .957,  .69,  .518}-0.21 & \cellcolor[rgb]{ 1,  .949,  .8}0.17 & 0.04 \\
anger & -0.01 & -0.02 & -0.05 & -0.02 & -0.01 & 0.00 \\
fear & 0.04 & 0.00 & -0.03 & 0.00 & -0.01 & -0.01 \\
surprise & \cellcolor[rgb]{ 1,  .949,  .8}-0.18 & 0.04 & -0.01 & -0.02 & 0.00 & -0.04 \\
\end{tabular}%
\end{table}
\begin{table}[b!]
\centering
\caption{Coefficients of Ridge Regression, predicting each $x$ variable from all final $z$ variables.
% \mom\ is compared with a machine-only benchmark (no side information) and with the human mapping applied to this machine model.
\label{tab:xonzcoef}}
\tblfontsz
\begin{tabular}{l|cccccc}
  & \feat{rate} & \feat{term} & \feat{dt} & \feat{rec} & \feat{inc} & \feat{emp} \bigstrut[b]\\
\hline
happiness & -0.07 & \cellcolor[rgb]{ .957,  .69,  .518}-0.29 & -0.10 & -0.06 & \cellcolor[rgb]{ .957,  .69,  .518}0.21 & -0.07 \bigstrut[t]\\
sadness & \cellcolor[rgb]{ 1,  .949,  .8}0.16 & 0.07 & 0.07 & -0.01 & \cellcolor[rgb]{ 1,  .949,  .8}0.13 & 0.07 \\
trustworthiness & \cellcolor[rgb]{ .514,  .235,  .047}\textcolor[rgb]{ 1,  1,  1}{-0.62} & \cellcolor[rgb]{ .957,  .69,  .518}-0.28 & -0.05 & \cellcolor[rgb]{ .957,  .69,  .518}-0.23 & \cellcolor[rgb]{ .514,  .235,  .047}\textcolor[rgb]{ 1,  1,  1}{0.31} & \cellcolor[rgb]{ 1,  .949,  .8}0.16 \\
dominance & 0.05 & \cellcolor[rgb]{ 1,  .949,  .8}0.16 & \cellcolor[rgb]{ 1,  .949,  .8}0.12 & \cellcolor[rgb]{ 1,  .949,  .8}-0.13 & -0.02 & 0.04 \\
hue & \cellcolor[rgb]{ .957,  .69,  .518}0.27 & \cellcolor[rgb]{ 1,  .949,  .8}0.20 & \cellcolor[rgb]{ 1,  .949,  .8}0.19 & 0.03 & 0.01 & -0.08 \\
eye gaze & \cellcolor[rgb]{ 1,  .949,  .8}0.13 & \cellcolor[rgb]{ .957,  .69,  .518}0.28 & -0.10 & \cellcolor[rgb]{ 1,  .949,  .8}0.13 & \cellcolor[rgb]{ .957,  .69,  .518}-0.29 & -0.04 \\
age & -0.09 & \cellcolor[rgb]{ 1,  .949,  .8}0.14 & \cellcolor[rgb]{ 1,  .949,  .8}0.12 & -0.09 & \cellcolor[rgb]{ .514,  .235,  .047}\textcolor[rgb]{ 1,  1,  1}{0.67} & \cellcolor[rgb]{ .514,  .235,  .047}\textcolor[rgb]{ 1,  1,  1}{0.40} \\
anger & 0.00 & 0.00 & 0.00 & 0.00 & 0.00 & 0.00 \\
fear & \cellcolor[rgb]{ 1,  .949,  .8}0.19 & \cellcolor[rgb]{ 1,  .949,  .8}0.12 & 0.08 & -0.07 & 0.04 & 0.00 \\
surprise & 0.07 & \cellcolor[rgb]{ 1,  .949,  .8}0.12 & 0.03 & -0.07 & -0.06 & \cellcolor[rgb]{ 1,  .949,  .8}0.13 \\
\end{tabular}%
\end{table}
%------------------------------------------------------------------------

%------------------------------------------------
%------------------------------------------------

\subsubsection{Learning Architecture.}

The network $\repnet$ contains a single linear layer with no bias
which takes a constant (1) as an input and outputs
% having no inputs and outputting
a number $z_i$ for each data dimension $i$.

The network $\hhat$ takes as input $(x,w,y)$. It contains one linear layer with no bias which takes as input $[x,y]$ and outputs a single number $\hat{s}$. The second linear layer (with bias) takes as input $w$ and outputs the sigmoid activation of a single number, $switch$, representing the propensity to incorporate $s$ at $w$. It then outputs $ w^{\intercal}x + switch \cdot \hat{s}$.

\subsubsection{Baselines.} 
\begin{itemize}
\item
\textbf{Machine Only}: %In the machine only baseline, we calculate the accuracy of
The best possible linear model (with bias) trained to predict $y$ from $x_1 \hdots x_{4}$. 
\item
\textbf{$h($Machine)}: %In the machine only baseline, we calculate the accuracy of
The human model $h$ applied to the best possible linear model (with bias) trained to predict $y$ from $x_1 \hdots x_{4}$. 
\[
h(\text{Machine}) = \beta_0 + h(x, \beta_1, \dots, \beta_4, s)
\]
where $\beta$ are the coefficients selected by the machine-only regression.
%   \todo{what are conditions}
\end{itemize}

\subsubsection{Human Models} 
\begin{itemize}
\item
\textbf{Always}: The human always fully incorporates the side information, 

$$h(x,w,s) = w^{\intercal}x + s$$
\item
\textbf{Never}: The human never incorporates the side information, 

$$h(x,w,s) = w^{\intercal}x$$
\item
\textbf{Or}: The human becomes less likely to incorporate side information as weight is put on $x_i, x_r$, 
$$h(x,w,s) = w^{\intercal}x + \sigma(1 / \textrm{max}(\textrm{max}(x_i,x_r),.0001) - 2). \cdot s$$
Note that max(.0001) is required to prevent numerical overflow, and -2 recenters the sigmoid to allow for values $<.5$.
\item
\textbf{Coarse}: The human incorporates $s$ as in Or, but uses a coarse, noisy version of $s$, $s' = 2 \cdot \1{s \geq 2}$
$$h(x,w,s) = w^{\intercal}x + \sigma(1 / \textrm{max}(\textrm{max}(x_i,x_r),.0001) - 2). \cdot s'$$
\end{itemize}

\section{Select Turker quotes}

\begin{itemize}
    \item
    ``I wasn't always looking at just happiness or sadness. Sometimes the expressions seemed disingenuously happy, and that also threw me off. I don't know if that was intentional but it definitely effected my gut feeling and how I chose.''
	\item
	``In my opinion, the level of happiness or sadness, the degree of a smile or a frown, was used to represent applications who were likely to be payed back. The more happy one looks, the better the chances of the client paying the loan off (or at least what the survey information lead me to believe).''
	\item
	``I was more comfortable with facial expressions than numbers. I felt like a computer and I didn't feel human anymore. Didn't like it at all.''
\end{itemize}
% \clearpage

% \includepdf{avatars_nips19.pdf}

\end{appendices}


%%%%%%%%%%%%%%%%%%%%%%%%%%%%%%%%%%%%%%%%%%%%%%%%%%%%%%%%%%%%%%%%%%%%%%%%%%%%%%%
%%%%%%%%%%%%%%%%%%%%%%%%%%%%%%%%%%%%%%%%%%%%%%%%%%%%%%%%%%%%%%%%%%%%%%%%%%%%%%%

\end{document}